\documentclass[10pt, a4paper]{article}

\usepackage{lrec-coling2024} 

\usepackage{graphicx}
\usepackage{booktabs}
\usepackage{tabularx}
\usepackage{soul}
\usepackage{xcolor}
\usepackage{linguex}
\usepackage{lscape}
\usepackage{ocgx}
\usepackage{tikz-dependency}
\usepackage[inline]{enumitem}	

\usepackage{hyperref}
\definecolor{darkblue}{rgb}{0, 0, 0.5}
\hypersetup{colorlinks=true, citecolor=darkblue, linkcolor=darkblue, urlcolor=darkblue}
\usepackage{xstring}
\usepackage{color}
\usepackage{subcaption}
\usepackage{multirow}
\usepackage{makecell}
\usepackage{tabularray}
\usepackage{tablefootnote}
\usepackage{float}
\usepackage[most]{tcolorbox}
\usepackage[whole]{bxcjkjatype}
\usepackage{cleveref}

\definecolor{seabornOrange}{HTML}{DE8F05}
\definecolor{fuchsia}{HTML}{966FD6}
\definecolor{asparagus}{HTML}{006B3C}
\definecolor{ashgrey}{HTML}{B2BEB5}
\definecolor{babyblue}{HTML}{A1CAF1}
\definecolor{banana}{HTML}{FFA812}
\definecolor{bittersweet}{HTML}{FE6F5E}
\definecolor{brass}{HTML}{B5A642}
\definecolor{sienna}{HTML}{E75480}
\definecolor{burlywood}{HTML}{DEB887}
\definecolor{celadon}{HTML}{ACE1AF}
\definecolor{chestnut}{HTML}{CD5C5C}
\definecolor{cyan}{HTML}{008B8B}
\definecolor{dollarbill}{HTML}{85BB65}
\definecolor{heliotrope}{HTML}{DF73FF}
\definecolor{indianyellow}{HTML}{E3A857}
\definecolor{airforceblue}{HTML}{5D8AA8}

\newcommand{\drugStop}[1]{\colorbox{brass}{#1$_{\texttt{drug}}^{\textcolor{red}{stopped}}$}\ }
\newcommand{\drugStart}[1]{\colorbox{brass}{#1$_{\texttt{drug}}^{\textcolor{red}{started}}$}\ }

\newcommand{\drugDec}[1]{\colorbox{brass}{#1$_{\texttt{drug}}^{\textcolor{red}{decreased}}$}\ }

\newcommand{\disorder}[1]{\colorbox{fuchsia}{\textcolor{white}{#1$_{\texttt{disorder}}$}}\ }

\newcommand{\timexPIT}[2]{\colorbox{ashgrey}{#1$_{\texttt{time}}^{\textcolor{red}{rel.~point~in~time}}$}\ }

\newcommand{\trigger}[1]{\colorbox{celadon}{#1$_{\texttt{change\_trigger}}$}\ }

\newcommand{\causedRel}[2]{\depedge[edge unit distance=0.8ex, label style={fill=yellow, font= \large}]{#1}{#2}{\texttt{caused}}}

\newcommand{\drawRelations}[2]{
    \begin{dependency}[theme=default]
        \begin{deptext}
            #1 \\   
        \end{deptext}
        #2          
    \end{dependency}
}

\title{A Dataset for Pharmacovigilance in German, French, and Japanese:\\Annotating Adverse Drug Reactions across Languages }

\name{Lisa Raithel$^{1,2,3,4\ast}$, Hui-Syuan Yeh$^{4,\ast}$\thanks{$^{\ast}$Shared first authorship; Corresponding author: raithel@tu-berlin.de},\\
      {\bf \large Shuntaro Yada$^5$, Cyril Grouin$^4$, Thomas Lavergne$^4$,}\\
      {\bf \large Aurélie Névéol$^4$, Patrick Paroubek$^4$, Philippe Thomas$^3$,}\\
      {\bf \large Tomohiro Nishiyama$^5$, Sebastian Möller$^{1,2,3}$, Eiji Aramaki$^5$,}\\
      {\bf \large Yuji Matsumoto$^6$, Roland Roller$^3$, Pierre Zweigenbaum$^4$}}

\address{$^1$BIFOLD, Ernst-Reuter Platz 7, 10587 Berlin, Germany;\\
         $^2$Quality \& Usability Lab, TU Berlin, Ernst-Reuter Platz 7, 10587 Berlin, Germany;\\
         $^3$German Research Center for Artificial Intelligence (DFKI), Alt-Moabit 91c, 10559 Berlin, Germany;\\
         $^4$Université Paris-Saclay, CNRS, LISN, Rue du Belvédère, 91405 - Orsay, France;\\
         $^5$Nara Institute of Science and Technology, 8916-5 Takayama-cho, Ikoma, Nara 630-0192, Japan;\\
         $^6$RIKEN, Nihonbashi 1-chome Mitsui Building, 1-4-1 Nihonbashi, Chuo-ku, Tokyo 103-0027, Japan
         }

\abstract{
User-generated data sources have gained significance in uncovering Adverse Drug Reactions (ADRs), with an increasing number of discussions occurring in the digital world. 
However, the existing clinical corpora predominantly revolve around scientific articles in English. 
This work presents a multilingual corpus of texts concerning ADRs gathered from diverse sources, including patient fora, social media, and clinical reports in German, French, and Japanese. 
Our corpus contains annotations covering 12 entity types, four attribute types, and 13 relation types. 
It contributes to the development of real-world multilingual language models for healthcare. 
We provide statistics to highlight certain challenges associated with the corpus and conduct preliminary experiments resulting in strong baselines for extracting entities and relations between these entities, both within and across languages.\\
\newline 
\Keywords{biomedical NLP, information extraction, adverse drug reactions, multilingual} }

\begin{document}

\maketitleabstract

\section{Introduction}

An adverse drug reaction (ADR) is a ``harmful or unpleasant reaction, resulting from an intervention related to the use of a medicinal product'' \citep{edwards_adverse_2000-1}.
ADRs constitute a significant problem in pharmacovigilance. 
No medication is without side effects, and even though there are clinical trials for each drug, the pool of patients included in the trials can never represent an entire population with respect to, e.g., age, gender, health, or ethnicity \citep{hazell_under-reporting_2006}. 
Even post-release surveillance campaigns might fail to reach the patients who have issues with the released medication \citep{hazell_under-reporting_2006}.
Therefore, medication use and effects must be monitored constantly.

Consequently, biomedical and clinical texts are a much-used resource for supporting pharmacovigilance since they contain information about patients, their medication intake, and, potentially, their medical history.
For example, researchers extract information from electronic health records (EHRs), scientific publications, public health or treatment guidelines, search logs \citep{white_early_2016}, and any other text dealing with medical issues.
However, all of these are written from the physician's perspective of treating the patient.

In contrast, social media, such as X (formerly Twitter) or Facebook, are created from the patient's perspective. 
Taken collectively, social media content can provide population-level signals for ADRs and other health-related topics.
Internet and social media engage many users and offer the means to access data at scale for specific topics of interest. 
Previous studies have shown that despite the large online user communities, they are not necessarily a representative sample of the population at large \citep{doi:10.1080/13691180801946150,wagner2015s}.  Nonetheless, people can use social media to anonymously discuss health issues in their own words without the fear of not being taken seriously, which is, in fact, one of the reasons for the under-reporting of ADRs, alongside a general mistrust of clinical providers \citep{yang_social_2012,palleria_limitations_2013}. 
Another factor making social media useful for detecting ADRs is the variety of languages provided on the internet, making health-related information more accessible to laypeople.
They, therefore, often turn to patient fora to research and collect information on topics they are concerned with, following ``translations'' from technical terminology to lay language provided by other members of the respective communities. 
Sometimes, there are even clinicians involved in these fora. 
This, again, highlights the necessity to extract relevant information not only from texts written by experts but also to listen to the patients' voices and process texts written by ``normal'' people.

Although the number of non-English and multilingual datasets is rising in the clinical and biomedical domain \citep{neveol_clinical_2018}, there is still a need for shareable corpora for particular tasks.
Especially the detection of ADRs, which is important across all countries and, therefore, languages, still shows much room for improvement, even in English data \citep{magge_deepademiner_2021-1}, but more so in other languages \citeplanguageresource{klein_overview_2020, raithel_cross-lingual_2022}.
Furthermore, shifting the perspective to the patient with the help of social media might support clinicians and other practitioners to understand their patients and the experienced ADRs better, react more appropriately, and meet the patients' needs more precisely \citeplanguageresource{arase_annotation_2020}. 
This also allows patients to participate actively in their treatment \citeplanguageresource{segura-bedmar_detecting_2014}.
Finally, the collected information from these crowd signals can be used for drug re-purposing and the development of new medications \citep{scaboro_increasing_2022}.
Therefore, with the presented work, we aim to broaden the access to resources for pharmacovigilance across languages and switch the perspective on health issues to the one of the patients.
We contribute to the development of real-world and multilingual models for patient-centric healthcare as follows:

\begin{itemize}
    \item We provide a new multilingual corpus focused on ADRs in three languages: German, French, and Japanese. 
    It is annotated with entities, attributes, and relations to describe experiences with ADRs from a patient's perspective.\footnote{\label{repo_url}Data and code can be found here: \url{https://github.com/Dotkat-dotcome/KEEPHA-ADR}}

    \item We describe the characteristics of the presented data and highlight challenges associated with the extraction of ADRs.
    
    \item We provide annotation guidelines, which aim to be robust across a variety of languages.\footnote{\url{https://github.com/DFKI-NLP/keepha_annotation_guidelines/blob/main/KEEPHA_annotation_guidelines.pdf}}

    \item We provide baseline models for named entity extraction, attribute classification, and relation extraction.\footref{repo_url}
\end{itemize}

\section{Related Work}

\subsection{Methods}
Since approximately 2010, with one of the first publications on the extraction of ADRs from social media by \citet{leaman_towards_2010-1}, the interest in and the number of social media datasets has been growing slowly.
By now, researchers, health-related industries, and authorities recognize the value of patient-generated data with respect to improving medication products and public health monitoring  \citep{sarker_utilizing_2015-1}.

Detecting and extracting ADRs from social media is done similarly to other information extraction tasks in NLP.
With the success of Transformer-based models \citep{vaswani_attention_2017-1} like \texttt{BERT} \citep{devlin_bert_2019} and \texttt{XLM-RoBERTa} \citep{conneau_unsupervised_2020} in almost all areas of NLP, these also started to dominate in the task of ADR extraction \citep{tutubalina_russian_2021, weissenbacher_overview_2022}.
Even so, there are still quite a few challenges in need of being addressed. 
The detection of ADRs in user-generated texts often suffers from small corpora (see \Cref{sec:datasets}), imbalanced label distributions, spelling mistakes, and colloquial language in general, and only a few language-specific medical Transformer models exist. 
Further, documents can contain ambiguous content and speculated statements or patients worrying about things that have not yet happened. 
These need to be distinguished from actual occurrences of ADRs. 

In the context of the Social Media Mining for Health\footnote{\url{https://healthlanguageprocessing.org/smm4h-2022/}} (SMM4H) 2022 shared task, \citet{portelli_ailab-udinesmm4h22_2022} address the limits of Transformers concerning document classification, entity extraction, and normalization.
They show that ensembling methods and architectures can improve the performance of these models, but also by using generative models like \texttt{GPT-2} \citep{radford_language_2019}.

\citet{miftahutdinov_kfu_2020} compare different model and data setups and demonstrate that a Convolutional Neural Network in combination with \texttt{fastText} embeddings \citep{bojanowski_enriching_2017-1} can outperform \texttt{mBERT} on Russian ADR texts in binary classification.
When using both English and Russian tweets for fine-tuning an English-Russian \texttt{BERT} model (\texttt{EnRuDRBERT}) they achieved the best result (within their experiments), especially when compared to only fine-tuning on Russian data. 
However, the authors also note that adding Russian data to the English data only improved the results on the English test set by one percentage point. 
\citet{gencoglu_sentence_2020} uses sentence embeddings \citep{reimers_sentence-bert_2019} to represent the tweets from SMM4H 2020, based on \texttt{RoBERTa} \citep{liu_roberta_2019-1} for English and multilingual \texttt{DistilBERT} \citep{sanh_distilbert_2019} for Russian and French to perform document classification.
They further weigh the contribution of positive samples to the loss function higher than the one of the negative samples to account for the label imbalance.  
With this, they achieve the best result within the shared task on the French dataset ($F_1 = 17\%$).

\citet{chowdhury_multi-task_2018} simultaneously classify posts and extract ADR and indication mentions from social media data in a multi-task setting.
They combine additive attention \citep{bahdanau_neural_2015} and a coverage mechanism \citep{see_get_2017} in a Recurrent Neural Network (RNN) and show that with this, mentions of ADRs are captured more accurately. 
\citet{raval_exploring_2021-2} model the tasks of ADR classification and extraction in a generative setting and use \texttt{T5} \citep{raffel_exploring_2020} for a sequence-to-sequence approach.
Adding temperature scaling \citep{devlin_bert_2019} and proportional mixing to account for different dataset sizes and languages improves performance in the binary classification of the French SMM4H 2020 dataset ($F_1 = 20\%$) compared with earlier results.

Similar work focuses on other types of text or medically relevant information. 
For example, \citet{meoni_large_2023} and \citet{agrawal_large_2022} study multilingual medical entity extraction using large language models (LLMs) with \texttt{InstructGPT} \citep{ouyang_training_2022} and \citet{feng_dkade_2023} propose \texttt{DKADE}, a framework incorporating a knowledge base that allows extracting ADRs and associated medication mentions in Chinese medical texts.

\subsection{Existing Datasets}\label{sec:datasets}

While there are many social media datasets related to the extraction of ADRs for English, e.g., the \textsc{CADEC} \citeplanguageresource{karimi_cadec_2015} and \textsc{PsyTar} \citeplanguageresource{zolnoori_systematic_2019} corpora, non-English datasets are rare.
We show those published in recent years in \Cref{tab:social_media_data} in \Cref{app:related_datasets} and describe them in more detail below. 

\paragraph{Spanish}

The \textsc{SpanishADR} corpus \citeplanguageresource{segura-bedmar_detecting_2014} was the first non-English social media dataset focused on ADRs.
The data originates from the patient forum ``ForumClinic''.
400 forum posts were randomly picked for annotation, and two annotators annotated adverse events and drug mentions.
\citetlanguageresource{segura-bedmar_detecting_2014} report an inter-annotator agreement (IAA) based on $F_1$ score of 0.89 for the drug mentions and 0.59 for adverse events.

\paragraph{Russian}
\citetlanguageresource{alimova_machine_2017} provided the first corpus in Russian.
They crawled the drug review forum Otzovik and created a corpus based on 580 reviews.
The reviews were annotated sentence-wise with one out of four labels: \textit{Indication, Beneficial effect, Adverse drug reaction, Other}.

\citetlanguageresource{tutubalina_russian_2021} created \textsc{RuDReC}, also originating from Otzovik.
The data is divided into two parts: one containing annotations on the entity level and the other one without annotations, comprising about 1.4 million texts from various online sources focused on health-related user posts. 
The annotated part comprises 500 documents.
The labels of the annotated corpus are sentence-based, marking whether or not health-related issues are mentioned using five different sentence labels.
Those that contain health problems were further annotated on the entity level, distinguishing six different entity types.
IAA was determined to be ``approximately 70\%'', using a relaxed agreement for entities following earlier work \citeplanguageresource{metke-jimenez_evaluation_2014,karimi_cadec_2015}.

\citetlanguageresource{sboev_analysis_2022} again harness reviews from Otzovik.
2,800 drug reviews are annotated with 18 entity types and additional attributes for drug and disease mentions.
Further, specific mentions are normalized to their respective concepts ICD-10 and MedDRA.
The accuracy achieved for the ADR entity type is 61.1\% using exact $F_1$ score.
    
\citetlanguageresource{klein_overview_2020} present a Twitter dataset made from Russian tweets with binary annotation.
The training set (the only one available) contains 7,612 tweets of which 666 describe an ADR.
For the test set, \citetlanguageresource{klein_overview_2020} list 1,903 tweets with 166 of those expressing an ADR.
The data was prepared for the fifth SMM4H shared task in 2020.

\paragraph{French}
For the same shared task, \citetlanguageresource{klein_overview_2020} further provide a French corpus based on data collected from Twitter.
The publicly available training set contains 2,426 tweets with 39 ADR examples.

\paragraph{Japanese}
\citetlanguageresource{arase_annotation_2020} published a corpus based on the Japanese patient forum TOBYO.
The authors crawled all entries related to lung cancer and containing one to five drugs from a pre-compiled dictionary.
The final corpus provides 169 documents annotated with drug effect spans, related drug mentions, types of reactions, and the ICD-10 codes for those. 
IAA was calculated using Fleiss' $\kappa$, resulting in $\kappa = 0.52$ for span and type agreement.

\paragraph{German}
Finally, the corpus provided by \citetlanguageresource{raithel_cross-lingual_2022} contains data from the German patient forum lifeline.de and is annotated with binary labels, expressing whether or not a document contains a mention of an ADR. 
Of the 4,169 documents, only 101 contain ADRs, showing a similar distribution of labels as other binary annotated corpora. 

\section{The Corpus}

Our new corpus contains data in three languages, German, French, and Japanese, based on the project collaborators' major languages.
The three languages belong to different families, and the data originates from different sources, representing diverse ways of expressing health-related issues written by laypeople.  
It is annotated with entities, attributes, and relations.
With this, we aim to capture relevant medical mentions from a patient's perspective.
We further add relationships between these entities to model, e.g., interactions between drugs and symptoms (that is, ADRs), body parts, or the patient's assessment of their well-being.  

\subsection{Data Collection}

The general requirements we set for the data were as follows:
\begin{enumerate*}[label=(\roman*)]
    \item The data should be health-related but not specific to any drug or disease,
    \item the data should be de-identifiable or already de-identified,
    \item the data should be distributable to other research teams.
\end{enumerate*}

\paragraph{German}

For the German data, we obtained permission from the administrators of the forum Lifeline\footnote{\url{https://fragen.lifeline.de/forum/}}, to download and share the data. 
On Lifeline, people discuss their experiences with specific diseases or medication and help each other in various life situations. 
We built a crawler and downloaded all posts available in the user forum in July 2021, containing posts between 2000 and 2021.
All messages were filtered for Covid-19-related posts, to remove potential vaccine-related reactions and discussions and avoid biasing our dataset towards this topic.
Ten thousand texts were randomly sampled and annotated with a binary label, expressing whether the text mentioned an ADR or not. 
Of these 10,000, 324 contained ADR mentions and were subsequently further annotated. 

\paragraph{French}

We found it very difficult to receive access to French patient fora.
For every potential resource, requirement (iii) would not have been met. 
Thus, we translated some of the already de-identified German texts and annotated them with binary labels to reduce the annotation and curation effort.
We used the DeepL\footnote{\url{https://www.deepl.com/translator}} machine translation service to translate German texts into French.
Then, we provided the texts to native French speakers who checked the texts for comprehensibility.
Minor issues were corrected by our annotators and texts that were not comprehensible due to an erroneous translation were discarded. 
Finally, we chose 100 translated documents containing ADRs for further annotation. 
Due to the relatively lower number of annotations in the French data compared to the other two languages, we designate French as a low-resource target language for our cross-lingual experiment in \Cref{subsec:experiment}. 
The set of French texts is distinct from the German texts to prevent data leakage in cross-lingual experiments.
See \Cref{app:details_de_fr} for more details on the French and German data.

\paragraph{Japanese}

The Japanese texts were collected from both Twitter and Yahoo! JAPAN Chiebukuro (YJQA)\footnote{\url{https://chiebukuro.yahoo.co.jp/}}, a Japanese Q\&A forum including health care issues.
For this, we had to relax requirement (i) for Twitter since searching for tweets without keywords is not possible due to the limitations of the Twitter API.
We collected tweets that mention the drug ``Lexapro'' and its ADR-related keywords (i.e., nausea, sleepiness, and appetite) from June 2017 to May 2020.
This drug is popular enough to be mentioned in social media and known as causing ADRs; we plan to extend the variety of drugs in future work.
For YJQA, we selected the Q\&As labeled ``concerns about ADR'' from an existing YJQA breast-cancer corpus \citeplanguageresource{kamba_medical_2021}, which is a labeled corpus of 1,000 randomly selected questions on breast cancer posted to YJQA from January 2018 to June 2020.
See \Cref{app:ja_data} for more details on the Japanese data.

\begin{table}[]
\centering
\small
\begin{tabular}{@{}p{.4\linewidth}p{.55\linewidth}@{}}
\toprule
\textbf{entity type}                        & \textbf{attributes} \\ \midrule

drug                               & increase, decrease, stopped, started, unique\_dose \\
change\_trigger                    & \\
disorder                           & negated\\
function                           & negated\\                
anatomy                            & \\              
test                               & \\          
opinion                            & positive, negative, neutral \\
measure                            & \\
time                               & frequency, duration, date, point in time \\
route                              & \\
doctor                             & \\
other                              & \\
user                               & \\
url                                & \\
personal\_info              & \\ \bottomrule
\end{tabular}
\caption{The different entity types and attributes. The bottom three are only for de-identification purposes. }
\label{tab:all_entities}
\end{table}

\begin{figure*}[th]
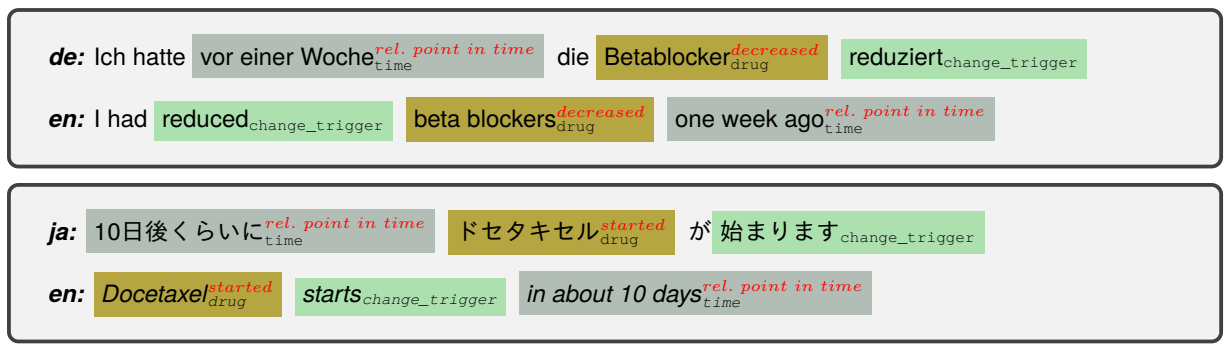

\small
    \begin{tcolorbox}    
    \textit{\textbf{de:}} Ich hatte \timexPIT{vor einer Woche}~ die \drugDec{Betablocker} \trigger{reduziert} \\ 
    
    \textit{\textbf{en:}} I had \trigger{reduced} \drugDec{beta blockers} \timexPIT{one week ago}
    
    \end{tcolorbox}

\footnotesize
    \begin{tcolorbox}
    \textit{\textbf{ja:}} \timexPIT{10日後くらいに}\\ \drugStart{ドセタキセル} が\trigger{始まります}\\
    
    \textit{\textbf{en:}} \textit{\drugStart{Docetaxel} \trigger{starts} \timexPIT{in about 10 days}.}
    \end{tcolorbox}
    \caption{Example annotation of a German (top) and Japanese (bottom) text, with their respective English translation.}
    \label{fig:ja_de_ner}
\end{figure*}

\subsection{Annotation}

The annotation guidelines were first developed using English examples from \textsc{CADEC} \citeplanguageresource{karimi_cadec_2015} and other English corpora related to ADRs. 
Annotation was conducted using \textsc{brat}\footnote{\url{https://brat.nlplab.org/}}.
After several pilot rounds of annotating these, the guidelines were applied to texts in the target languages and further refined. 
Then, our annotators, all (near-) native speakers of the respective languages\footnote{See \Cref{app:annotator_background} for more information on the annotation process and our annotators.}, were trained the same way, i.e., by first annotating English examples.
These annotations were subsequently discussed and any constructions that could not be modeled with our annotation scheme were further investigated to determine if these were language-agnostic or applied to one language only. 
Ultimately, we decided to annotate 12 entity types, four attribute types, and 13 relation types. 
They are shown in \Cref{tab:all_entities} and \Cref{tab:relations}.

\paragraph{De-identification}
For de-identifying private user information, such as names and addresses, we first tried to find as many identifiers as possible using regular expressions specific to the respective sub-dataset.
Very frequent occurrences were, for instance, user names, with which patients greet each other and/or refer to each other using nicknames (or nicknames of nicknames, for example, ``Mohnblümchen'' for the user name ``Mohnblume'', a diminutive of ``poppy'').
We collected the regular expression matches and replaced them with a mask (\texttt{<user>}) to keep the structure of the text intact.
However, since users are very creative in inventing names, greetings, and goodbyes, not all of them were captured. 
Therefore, one of the tasks for the annotators was to add an entity label \texttt{user} to all still-existing names during the entity annotation process.
Those were then replaced after the annotation was completed.
We did the same for URLs, (e-mail) addresses, and other potentially de-identifying information.

\paragraph{Entities}
We annotate entities in the form of noun- or verbal phrases together with their modifying parts, e.g., adjectives and adverbs.
Complex modifiers often occurring in Japanese are excluded to support the language-independence of the guidelines.  
We always prefer the smallest core noun phrase or, otherwise, the whole verb phrase.
Metaphors, descriptive language, and spelling mistakes are annotated as well. 
Discontinuous entities are allowed if necessary. 

In more detail, we annotate drug mentions (\texttt{drug}) and any description of medical signs or symptoms, no matter whether or not they are an ADR (\texttt{disorder}).
We further annotate trigger words or phrases that mark a change in medication intake (\texttt{change\_trigger}) as well as mentions describing any part of the body (\texttt{anatomy}).
Next, we mark body functions (\texttt{function}), i.e., normal processes of the body, like ``sleep'' or ``appetite'', which can sometimes be negated, similar to disorders. 
Medical tests (\texttt{test}) and resulting measurements or medication dosages (\texttt{measure}) are labeled, too, as well as the means of medication intake (\texttt{route}).
Further, we mark the assessment and evaluation of patients with respect to drugs, disorders, or functions using the entity type \texttt{opinion}. 
Since timelines are also an essential concept in medication intake, we apply a label called \texttt{time} to any mention expressing a time, e.g., a duration or a frequency. 
Finally, doctors' professions are labeled with \texttt{doctor}, and all remaining entities that seem relevant to the annotator can be marked with \texttt{other}.

\paragraph{Attributes}
Some entities are extended by attributes. 
For example, the \texttt{drug} entity can be further specialized by adding a marker that represents the current state of the drug, e.g., whether it was recently started or stopped by the patient (or by prescription). 
Mentions of body functions and disorders can be negated, for example, in the case the medication helped the patient and the described symptoms do not exist anymore. 
Patients' opinions on drugs or disorders can be attributed as positive, negative, or neutral. 
Lastly, time expressions can be marked as describing, e.g., a frequency or a duration.

\begin{table}[h]
\centering
\footnotesize
\begin{tabular}{@{}lp{.25\linewidth}p{.25\linewidth}@{}}
\toprule
\textbf{relation type}          & \textbf{head}                      & \textbf{tail}\\ \midrule
caused                          & drug, disorder                 & disorder, function \\
treatment\_for                  & drug                           & disorder, function \\
has\_dosage                     & drug                           & measure \\
experienced\_in                 & disorder                       & anatomy \\
examined\_with                  & disorder, anatomy, function    & test \\
has\_result                     & test                           & measure, disorder, function \\
refers\_to                      & disorder                       & disorder, function$^{\textsl{negated}}$ \\
refers\_to                      & drug                           & drug \\
refers\_to                      & anatomy                        & anatomy \\
refers\_to                      & function                       & function \\
interacted\_with                & drug                           & drug \\
signals\_change\_of             & change\-trigger                & drug \\
has\_time                       & drug, disorder\                  & time \\
has\_route                      & drug                           & route \\
is\_opinion\_about              & opinion                        & drug, disorder, function \\
misc                            & ANY                            & ANY \\ \bottomrule
\end{tabular}
\caption{Overview of available relation types and the entity types they associate. 'ANY' stands for any entities we defined.}
\label{tab:relations}
\end{table}

\paragraph{Relations}

\begin{figure*}[th]
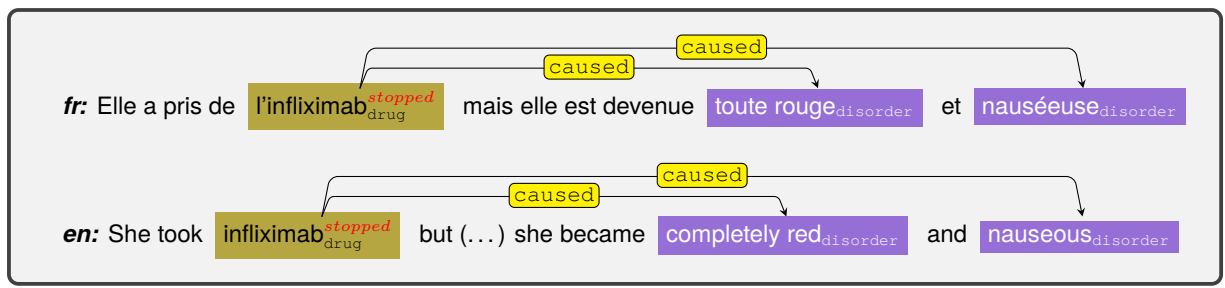

\small
    \begin{tcolorbox}
        \drawRelations{\textbf{\textit{fr:}}
        Elle a pris de \& \drugStop{l'infliximab} \& mais elle est devenue \& \disorder{toute rouge} \& et \& \disorder{nauséeuse}}
        {
            \causedRel{2}{4} 
            \causedRel{2}{6}
            }\\
        
        \drawRelations{\textbf{\textit{en:}}
        She took \& \drugStop{infliximab} \& but (\ldots) she became \& \disorder{completely red} \& and \& \disorder{nauseous}}
        {
            \causedRel{2}{4} 
            \causedRel{2}{6}
            }
            \end{tcolorbox}
\caption{An example annotation of a \texttt{caused} relation taken from the French dataset and translated. According to the writer of this message (patient), the medication \textit{infliximab} is likely to have caused the symptoms \textit{toute rouge} and \textit{nauséeuse}.}

\label{fig:fr_rel}
\end{figure*}

The most important relation type is \texttt{caused} and differentiates our corpus from those of other work described in \Cref{sec:datasets}: 
We do not specifically mark mentions of ADRs with an ``ADR'' label but only express ADRs with the \texttt{caused} relation between medications and symptoms (or body functions). 
This relation type can also be used to mark disorders that are the reason for other disorders or body functions. 
Further, we represent treatments of medical signs or symptoms with the \texttt{treatment\_for} relation.
Medications and their routes and dosages can be connected via the types \texttt{has\_route} and \texttt{has\_dosage}, respectively.
To connect medical symptoms with a body part, we introduce the relation type \texttt{experienced\_in}.
Moreover, disorders, body parts, and functions can be \texttt{examined\_with} a test.
Those tests can have results (\texttt{has\_result}) in the form of measures (like a certain cholesterol value), but also in the form of diagnoses, expressed as disorders or functions.
In case there is evidence that a medication interacted with another one, this can be modeled using the \texttt{interacted\_with} relation type. 
Triggers of medication change can be associated with the \texttt{signals\_change\_of} relation to a drug mention.
Furthermore, drugs and disorders can be connected to time expressions with the \texttt{has\_time} relation, to mark the time of medication intake or the duration of a symptom. 
To represent assessments by patients concerning drugs, disorders, and functions, we introduce the \texttt{is\_opinion\_about} type.
Finally, we add a \texttt{refers\_to} relation type to connect co-referring mentions, e.g., in case patients first mention a medication name and afterward only an abbreviation of it. 
All associations that seem relevant to the annotators but are not represented in our annotation scheme can be modeled with the \texttt{misc} relation.
We \emph{do not} annotate relations if the relevant entities are part of a hypothetical or speculative statement or a question. 
See examples of annotated texts in \Cref{fig:ja_de_ner} and \Cref{fig:fr_rel}.

\subsection{Final Dataset}

In total, the corpus contains 118 texts in German with a minimum of 55 tokens per text, 100 texts in French with at least 42 tokens per text, and 619 texts in Japanese, with the shortest text containing 15 characters.
See \Cref{fig:text_len_distribution} for the distribution text lengths per language.
The number of entities, relations, and attributes per language is shown in \Cref{tab:overview_detailed_annos}, and more detailed statistics are shown in \Cref{tab:overview_final_dataset} (\Cref{app:data_stats}).
The German data were annotated by two annotators and subsequently consolidated, while the other two datasets were annotated by one annotator each.

\begin{table}[th]
\centering
\begin{tabular}{@{}lrrrr@{}}
\toprule
\textbf{language} & \textbf{\#doc} & \textbf{\#ent} & \textbf{\#rel} & \textbf{\#attr}         \\ \midrule
de & 118                  & 3,487               & 2,163                & 1,141             \\
fr & 100                  & 1,939               & 1,129                & 537                \\ 
ja & 619                   & 9,464               & 5,083                & 2,364                   \\ 
\bottomrule
\end{tabular}
\caption{Number of documents (\#doc) with the number of entities (\#ent), relations (\#rel), and attributes (\#attr) of each type for each language (lang.). 
}
\label{tab:overview_detailed_annos}
\end{table}

\paragraph{German}
The by far most frequent entity type for the German part of the data is \texttt{disorder} (1,151 annotations), followed by \texttt{drug} (642 annotations). 
The entity type with the lowest frequency for German is \texttt{route}.
Furthermore, \texttt{caused} is the most often annotated relation type. For attributes, we find that the \texttt{time} attribute was used quite often by the annotators (622 times), mostly with \texttt{duration} or \texttt{point in time} as attribute values. 
The inter-annotator agreement for the German data was a micro average (relaxed) $F_1$ score of 0.77 for entities (with \texttt{drug} and \texttt{disorders} showing an agreement of 0.93 and 0.84, respectively), 0.38 for relation types (the annotators agreed on the \texttt{caused} relation with a score of 0.60) and 0.41 for attributes. 
Note that the relation and attribute annotation was conducted in the same session as the entity annotation, so disagreements in the entity annotation were propagated to the other layers. 

\paragraph{French}
For French, the most often annotated entity type is \texttt{disorder} (588 mentions).
Similar to the German data, it is followed by \texttt{drug}. 
\texttt{test} is the entity type with the fewest annotations.
The \texttt{caused} relation is annotated in the French data 342 times, showing the highest frequency.
Also, \texttt{has\_time} has the second-highest frequency for French.
Attributes for time expressions and medication mentions are used the most.

\paragraph{Japanese}
The distribution of the entities in the Japanese data follows the same pattern as German and French: \texttt{disorder} and \texttt{drug} are the most frequent types, with 2,843, and 2,031 occurrences, respectively. 
Doctors' names are listed last in terms of frequency.  
Regarding relations, we again see the same pattern as in the other two languages: \texttt{caused} and \texttt{has\_time} are the most frequent relation types with 979 and 753 annotations, respectively.
For Japanese, attributes for \texttt{opinion} and \texttt{time} occur most often. 

\begin{table}[th]
\small
\centering
\begin{tabular}{@{}p{.20\linewidth}p{.35\linewidth}p{.35\linewidth}@{}}
\toprule
\textbf{drug}    & \textbf{disorder (de)}               & \textbf{translation (en)}       \\ \midrule
ads              & Gelenk\-schmerzen                 & joint pain                 \\
estreva gel      & vermehrte, starke Kopfschmerzen & increased severe headaches \\
cerazette        & 3kilo runter                    & 3 kilos down               \\
opipramol        & Watte im Kopf                   & ``cotton in my head''          \\
mtx              & Haarausfall                     & hair loss                  \\
venaflaxin       & Unwirklich\-keits\-gefühle          & feelings of unreality      \\ 
utrogest         & wilde Träume                    & wild dreams                \\
\bottomrule
\end{tabular}
\caption{A random selection of ADRs (i.e., medication mentions that have caused disorders) extracted from the German part of the corpus. }
\label{tab:adr_de_short}
\end{table}

In summary, we can see similar patterns in the annotations across languages, e.g., the distribution of entity types.
However, the Japanese dataset is much larger than the German and French datasets, which is also evident in the number of entity and relation annotations in total.
Apart from being less distant languages, the French and German data originate from the same source, and this also shows in the distribution of annotations. 

Due to the colloquial style of the text, it was sometimes difficult for the annotators to pinpoint the boundaries of entities since some descriptions are rather ``creative'' or even metaphorical (see \Cref{tab:adr_de_short}). 
Similarly, finding the exact attribute for, e.g., a medication mention, was challenging as well since patients often describe starting and stopping their medication intake in one and the same sentence. 
For illustration, we show some extracted information from the German part of the corpus in \Cref{tab:adr_de_short}, together with their translations. 
Note that these extracted phrases still need to be verified by a pharmacovigilance expert.

\section{Experiments}

The following describes our baseline models, which will be published together with the data and are meant to serve as a starting point for future work.
We show experiments on named entity recognition (NER), attribute classification (AC), and relation extraction (RE).

We evaluate the performance of the models detailed below using \textsc{brat} format to account for span boundaries independently of the tokenizer.
Therefore, for NER, we convert the predictions of the models back to \textsc{brat} and evaluate them using ``brat eval''\footnote{
\url{https://perso.limsi.fr/pz/blah2015/}.}.
We then report micro and macro average $F_1$ scores for all tasks, calculated on relaxed boundaries for NER.

\subsection{Task Setup}
We carried out the three tasks independently. 
The hyper-parameter tuning is performed for each dataset combination for each task.

\paragraph{Name Entity Recognition (NER):} The dataset includes discontinuous and overlapping entity annotations (see \Cref{tab:complicated_anno} in \Cref{app:data_stats} for details), and preparing these annotations for model fine-tuning requires complex methods \citep{baldini_soares_matching_2019,dai_effective_2020-2,dirkson_fuzzybio_2021-1,li_span-based_2021}. 
Since these special entity annotations are infrequent, we remove sentences containing them during model fine-tuning. 
This helps prevent models from encountering and potentially struggling with special annotations without employing more complex handling methods. 
We convert the \textsc{brat} annotations to \emph{BILOU} format \citep{ratinov_design_2009} for fine-tuning. 
For evaluation, we convert the predicted \emph{BILOU} tags back to \textsc{brat} format.

\paragraph{Attribute Classification (AC):} 
Similar to RE, we extract the sentences covering the entity corresponding to the attribute and use the special token pair \texttt{[E]}, \texttt{[/E]} to mark the entity.

\paragraph{Relation Extraction (RE):} 
To prepare each relation sample from the document-level annotations, we extract only the sentences containing entities from the documents. 
We use special token pairs \texttt{[E1]}, \texttt{[/E1]} and \texttt{[E2]}, \texttt{[/E2]} \citep{baldini_soares_matching_2019} to enclose the head and tail entities.

\subsection{Experiment Setup}
\label{subsec:experiment}
For all experiments, we fine-tune \texttt{XLM-RoBERTa}$_{large}$ \citep{conneau_unsupervised_2020}, henceforth \texttt{XLM-R}, on the respective downstream task.
The model supports French, German, and Japanese, among other languages. 
The different settings are aimed at investigating the performance within and across languages. 

\paragraph{Mono-lingual:} 
We fine-tune and test \texttt{XLM-R} on each language of the dataset separately, for French, German, and Japanese, respectively.

\paragraph{Multi-lingual:} 
We mix the languages while fine-tuning; 
each batch samples from each language proportionally to the size of this language in the training set.
The fine-tuned multilingual models are evaluated on each language separately and across languages.

\paragraph{Cross-lingual:} 
We apply the model in a zero-shot cross-lingual transfer setting, i.e., we (1) fine-tune \texttt{XLM-R} on the source language(s) and (2) directly test the model on the target languages.

\begin{table*}[t]
\centering
\footnotesize
\begin{tabular}{l cc ccc cc cc}
\toprule
\multirow{2}{*}{\textbf{Experiments}} & \multirow{2}{*}{\textbf{\makecell{\emph{train}}}} & \multirow{2}{*}{\textbf{\makecell{\emph{test}}}} &  \multicolumn{2}{c}{\textbf{NER (\%)}} &  \multicolumn{2}{c}{\textbf{AC (\%)}} & \multicolumn{2}{c}{\textbf{RE (\%)}} \\
\cmidrule(r){4-5}
\cmidrule(r){6-7}
\cmidrule{8-9}
 & & &  \textbf{micro F1} & \textbf{macro F1} &  \textbf{micro F1} & \textbf{macro F1} & \textbf{micro F1} & \textbf{macro F1} \\
\hline
\addlinespace[4pt] 
\multirow{3}{*}{Mono-lingual} 
& de & de & 75.8  & 65.4 & 76.8 & 56.9 & 79.3 & 75.7 \\ 
& fr & fr & 82.5  & 71.9 & 84.4 & 73.8 & 87.0 & 78.2 \\ 
& ja & ja &  61.0 & 58.5 & 85.8 & 81.0 & 87.2 & 80.4 \\

\midrule
\multirow{4}{*}{Multi-lingual} 
& de+fr+ja &  de        & 77.3 & 67.6  & 80.4 & 66.9 & 83.4 & 79.2 \\
& de+fr+ja &  fr        & 83.9 & 75.3  & 90.8 & 82.8 & 88.3 & 82.0  \\
& de+fr+ja &  ja        & 64.5 & 65.1  & 88.0 & 82.6 & 86.5 & 78.0 \\
& de+fr+ja &  de+fr+ja  & 74.1 & 69.3  & 85.8 & 71.7 & 85.9 & 76.7  \\

\midrule 
\multirow{3}{*}{Cross-lingual}  
& de & fr & 77.3 & 68.8 & 69.5 & 63.6 & 78.7 & 79.3 \\
& de & ja & 48.8 & 38.8 & 53.7 & 41.3 & 62.2 & 54.5\\
& de+ja & fr & 77.5 & 66.7 & 80.8 & 71.2 & 83.2 & 75.9 \\

\bottomrule
\end{tabular}
\caption{Average scores of models fine-tuned on five different seeds on the KEEPHA dataset with different language combinations. The underlying pre-trained model for all experiments was \texttt{XLM-RoBERTa}$_{large}$.}
\label{tab:results}

\end{table*}

\subsection{Results}
The results are described in the following and shown in \Cref{tab:results}.

\paragraph{Mono-lingual:}
Regarding the results for AC and RE, we see that the \textbf{ja} models perform the best and are closely followed by \textbf{fr} with \textbf{de} being last with a larger difference in scores. 
This, in turn, follows the pre-training data size of \texttt{XLM-R}\footnote{Based on the CommonCrawl Corpus \citeplanguageresource{wenzek_ccnet_2020}, the order in terms of data size is \textbf{ja} (69.3 GiB) $>$ \textbf{de} (66.6 GiB) $>$ \textbf{fr} (56.8 GiB).}.
However, for NER, the performance is \textbf{fr} $>$ \textbf{de} $>$ \textbf{ja}, with \textbf{ja} falling to the last place.

\paragraph{Multi-lingual:}
In general, the multilingual models fine-tuned on all languages boost performance across all tasks and languages, except for \textbf{ja} in RE. 
When comparing with the monolingual experiments, \textbf{fr} outperforms marginally in AC and RE, benefiting from the contributions of the other two languages.

\paragraph{Cross-lingual:}
We observe that the models fine-tuned on \textbf{de} and evaluated on \textbf{fr} work well and only show a modest decrease from the monolingual models trained on \textbf{fr} ($-1\%$ for RE; $-3\%$ for NER; $-10\%$ for AC macro $F_1$). 
The models fine-tuned on \textbf{de} and evaluated on \textbf{ja} are still far behind the monolingual model trained on \textbf{ja} only ($-20\sim30\%$ macro $F_1$). 
When comparing models fine-tuned on \textbf{de+ja} and evaluated on \textbf{fr} to models fine-tuned on \textbf{de} only, we observe consistent improvements in micro $F_1$ across the three tasks, but a drop in macro $F_1$ for NER and RE.

\section{Discussion \& Conclusion}

With this work, we provide a new corpus of texts in German, French, and Japanese to support pharmacovigilance across languages by extracting information on ADRs from user-generated content. 
Training models on this corpus might facilitate information aggregation across countries, which is important for detecting rare diseases or adverse reactions. 
Furthermore, gathering and analyzing data globally can help develop new medications or treatments and benefit minorities. 
The corpus is annotated based on annotation guidelines carefully designed to apply to German, French, Japanese, and also English, potentially allowing the guidelines to be used for other languages as well. 
Annotations are conducted on entity, attribute, and relation levels to cover as much information as possible. 
By choosing languages from different language families and cultures, we provide a challenging resource with which we hope to advance the detection of ADRs and other medically relevant expressions.

To initiate future work, we provide baseline models for all three tasks, i.e., named entity recognition, attribute classification, and relation extraction, highlighting the difficulties of state-of-the-art Transformer models when faced with complex domain-specific data. 

Further future work will focus on improving the cross-lingual performance of available models, for example in combination with few-shot approaches and/or large language models, such as \texttt{Llama} \citep{touvron_llama_2023-1}.
A more detailed investigation into the impact of the different data sources on the overall performance of the models might further deepen the understanding of the data, too.
Moreover, investigating cross-cultural differences in how people discuss their health issues online is an exciting topic to explore. 
Building on the work of \citet{scaboro_increasing_2022}, who analyzed negation and speculation constructions, examining specific syntactic structures and linguistic phenomena, as well as potential biases in the new corpus would be interesting, too.
Finally, we are already working on extending our corpus with more data. 
For instance, we are annotating Japanese case reports and social media messages containing a more diverse pool of medications.
Moreover, we aim to gain access to more (original) French data to diversify the dataset even more.
Normalizing disease descriptions to medical ontologies will be one of the next steps as well. 

\section{Ethical Considerations}

When using data from social media, we commit to a particular sub-group of people: 
Those who have access to and actively participate on these platforms. 
Depending on the platform, the age range might vary, too.
Again, this introduces a bias, which can be learned by the respective language models fine-tuned on these data. 

Also, the presented dataset is only a small glimpse of ADR-related topics discussed online.
The German and French parts of the corpus, particularly, are very similar due to the translation.
More different sources and languages need to be considered to make the dataset more diverse. 

Several ethical aspects need to be considered when creating the dataset. 
First, the de-identification might not be perfect, i.e., even if usernames, etc., are masked, it might still be possible to identify the users since the fora are publicly accessible.
The corpus will only be distributed via a data protection agreement and only within the research community. 
Second, the extracted information should not be further processed as is but instead verified by a pharmacovigilance expert. 
One mention of a potential ADR in a user post does not make an ADR per se, but this information should be further investigated. 
Related to this, normalizing user descriptions to medical ontologies would also make it easier for experts to analyze potential health risks. 
Also, the automatic translation of the German texts into French might have introduced some biases.

Regarding the language model we used to conduct the baseline experiments (\texttt{XLM-RoBERTa}$_{large}$), we cannot rule out the existence of sensitive contents in the pre-training data, which might also have introduced biases into the models.

\section{Acknowledgements}

First and foremost, a heartfelt `thank you' to our annotators Alon Drobickij, Selin Yeginer, Garance Forestier, Emiliano Valdes Menchaca, and Narumi Tokunaga, for doing amazing work.
We further thank the anonymous reviewers for their constructive feedback on this paper. 
Our work was supported by the Cross-ministerial Strategic Innovation Promotion Program (SIP) on ``Integrated Health Care System'' Grant Number JPJ012425, by JPMJCR20G9, ANR-20-IADJ-0005-01, and DFG-442445488 under the trilateral ANR-DFG-JST AI Research project KEEPHA, and by the German Federal Ministry of Education and Research under the grant BIFOLD24B.

\section{Bibliographical References}\label{sec:reference}

\bibliographystyle{lrec-coling2024-natbib}
\bibliography{lrec2024}

\section{Language Resource References}
\label{lr:ref}
\bibliographystylelanguageresource{lrec-coling2024-natbib}
\bibliographylanguageresource{lrec2024}

\onecolumn
\appendix

\section{Related Datasets}\label{app:related_datasets}

Other non-English social media corpora focused on ADRs. 
The number of documents (\#documents) refers to the definition of documents per corpus, i.e., some are sentence-based, some are post-based, etc. 
Some test sets are unavailable to the public since they are/were part of a shared task.

\begin{table}[ht]
   \centering
   \small
   \begin{tabular}{lrrrr}
   \toprule
       \textbf{language} & \textbf{\#documents} & \textbf{type}         & \textbf{annotation}              & \textbf{authors} \\ \midrule
       es       &  400   & forum        & entities                & \citetlanguageresource{segura-bedmar_detecting_2014}   \\ 
       ru       &  $^\ast$279   & drug reviews & multi-label             & \citetlanguageresource{alimova_machine_2017} \\
       fr       &  3,033 & Twitter      & binary                  & \citetlanguageresource{klein_overview_2020} \\
       ru       &  9,515 & Twitter      & binary, entities        & \citetlanguageresource{klein_overview_2020} \\
       ja       &  169   & forum        & entities, normalization & \citetlanguageresource{arase_annotation_2020}  \\
       ru       &  $^{\ast\ast}$500  & drug reviews & multi-label, entities   & \citetlanguageresource{tutubalina_russian_2021} \\
       ru       &  2,800 & drug reviews & entities                & \citetlanguageresource{sboev_analysis_2022} \\
       de       &  4,169 & forum        & binary                  & \citetlanguageresource{raithel_cross-lingual_2022} \\ \midrule
       de, fr, ja & 837 & forum, Twitter, YJQA & entities, attributes, relations & ours \\
       \bottomrule
   \end{tabular}
   \caption{Other non-English social media corpora focused on ADRs. es=Spanish, fr=French, ru=Russian, ja=Japanese, de=German. $^\ast$Number of documents containing ADRs. $^{\ast\ast}$This is only the annotated part of the \textsc{RuDReC} corpus.}
   \label{tab:social_media_data}
\end{table}

\section{Details on German and French Data} \label{app:details_de_fr}

To avoid potential confusion, we provide an explanation for the relation between the \textsc{Lifeline} corpus (German) provided by \citetlanguageresource{raithel_cross-lingual_2022} and the data presented in this paper. 

\paragraph{French}
The \textsc{Lifeline} corpus \citeplanguageresource{raithel_cross-lingual_2022} contains 4,169 documents in German, crawled from the patient forum lifeline.de.
These documents are labeled with binary classes, i.e., either a document mentions an adverse drug reaction or does not.
We took these documents and automatically translated them into French.
The translations were validated and improved (if necessary) by French speakers (our annotators). 
We then took the 100 positive documents that remained after validation (those containing mentions of ADRs) and prepared them for further annotation.
These documents, therefore, overlap with the positive documents in \citetlanguageresource{raithel_cross-lingual_2022} but are semi-automatically \textit{translated}.

\paragraph{German}
The German documents presented in this paper \textit{do not} overlap with the data in \citetlanguageresource{raithel_cross-lingual_2022}, but they are extracted from the same forum.

\paragraph{De-Identification}
Regarding the masking of identifying information, for French and German, the following applies: 
The German data is from a public anonymous forum, so they can be found already publicly on the web. 
Of course we have to acknowledge that often, people on any kind of social media are not necessarily aware of the potential reach of their posts. 
However, by masking details, we try to cut the connection between our documents and the original forum posts to make it more difficult to trace the documents back to their original.

As the French were a translation of part of the already de-identified German data, these documents were subject to the same de-identification procedure, plus an additional modification of the documents by translation. 
Even with a translation of the documents back to German, it is difficult to trace the documents back to their original.

\section{Details on Japanese Data} \label{app:ja_data}

The Japanese tweet documents contain 20 tweets per document, and each Q\&A document includes one question and one answer text, resulting in 99 full documents as shown in \Cref{tab:overview_detailed_annos}.
The definition of ``token'' or ``word'' in Japanese may change according to different Japanese grammar theories. 
We can estimate what space-delimited languages call `the number of ``words''' based on the average character counts in Japanese ``words'', which more-or-less span 2.5--3.5 characters.

\paragraph{De-Identification} 
We have two sources for Japanese: Twitter (currently X) and an online forum. 
We applied basic regular expressions to identify Twitter user names and URLs in our data. 
Then, we manually de-identified all potential mentions of private information, such as names of persons, hospitals, and organizations.

\section{Annotation Process and Annotators}\label{app:annotator_background}

\paragraph{Annotation Process}
For annotation, we used the widely known annotation tool \textsc{BRAT}, since almost everyone on the team was familiar with it, it allows the annotation of attributes, and it was furthermore favorably reviewed by \citet{neves_extensive_2021}.

The guidelines design and pilot annotation were mainly done using English data since we did not have many positive documents in Japanese, French, or German. 
For all languages, the first annotations were therefore conducted by the annotators and some of this paper's authors on English texts, with data taken, e.g., from the \textsc{CADEC} corpus \citeplanguageresource{karimi_cadec_2015}. 
We annotated examples in two rounds for three months and discussed/refined the entity and relation scopes.
When these pilot annotations reached a satisfactory IAA, we asked our annotators to label the data in the respective languages. 
During annotation, whenever problems occurred, the annotator discussed with the annotation instructors (the first and third author), which involved a small number of working-level modifications to the guidelines.
We generally followed the state-of-the-art methodology recommended by  \citet{fort_collaborative_2016}.

\paragraph{Annotators}
\Cref{tab:annotators_background} lists our annotators.
Each annotator except one (A5) in \Cref{tab:annotators_background} is an enrolled student and employed as a student assistant with varying working hours, depending on their availability. 
Annotators A1 to A4 earn(ed) 12,95€ per hour.
Annotators can distribute their working hours freely, with the recommendation to annotate not more than two hours continuously. 
Annotator A5 is a full-time employee who annotates Japanese corpora.

\begin{table*}[h]
\small
\centering
\begin{tabular}{@{}lp{.10\linewidth}p{.30\linewidth}p{.20\linewidth}p{.08\linewidth}p{.10\linewidth}@{}}
\toprule
\textbf{annotator} & \textbf{working language}  & \textbf{knowledge of languages}  & \textbf{study program}    & \textbf{entry}     & \textbf{working hours} \\ \midrule
A1                 & de, en                                                               & German and Russian bilingual, good knowledge of English                          & Pharmacy, Freie Universität Berlin, Germany     & November 2021                                                         & 10                                                               \\
A2                 & de, en                                                               & German and Turkish bilingual,  good knowledge in English,  basics in French and Spanish & Pharmacy, Freie Universität Berlin, Germany     & March 2022                                                            & 10                                                               \\
A3                 & fr, de                                                               & French (native), German (C1),  English (C1)                                      & Human Medicine, Charité Berlin, Germany            & May 2023                                                              & 8                                                                \\
A4                 & fr, de                                                               & Spanish (native), French (C2), German (C1), good knowledge of English           & Life Science Engineering,  HTW Berlin,  Germany & August 2023                                                           & 20                                                               \\
A5                 & ja                                                               & Japanese (native) with good knowledge of English                                       &  (MSc in Biomedicine)            & April 2021                                                              & full-time                                                                \\ \bottomrule
\end{tabular}
\caption{The background information of our annotators. 
The table shows the languages they were working on, the languages they know in general, their study programs (or obtained degree), the time they were hired, and their working hours per week.}
\label{tab:annotators_background}
\end{table*}

\section{Dataset Statistics}\label{app:data_stats}

\begin{figure}[h]
\centering
\begin{subfigure}[h]{0.42\linewidth}
\includegraphics[trim=0 0 0.5cm 0, clip, width=\linewidth]{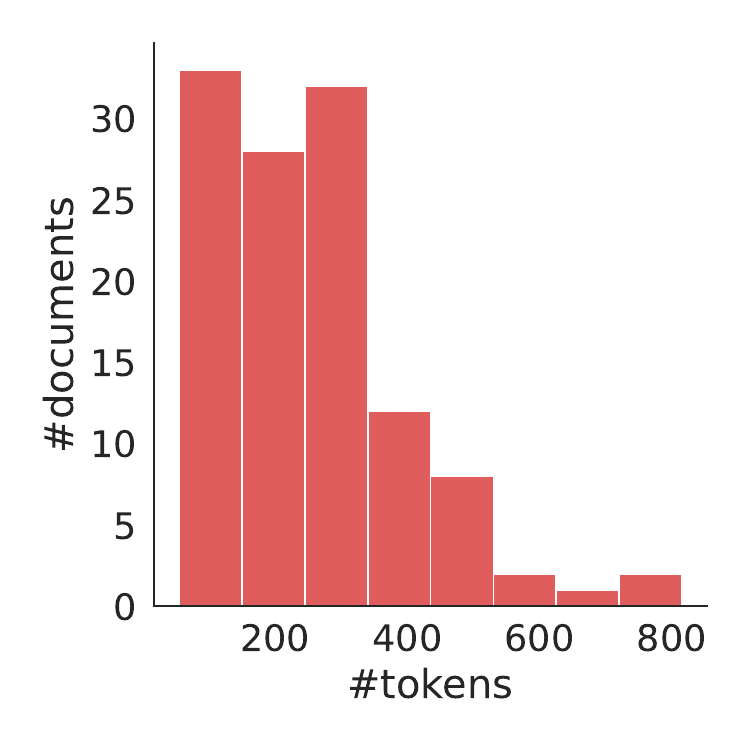}
\caption{German.}
\end{subfigure}%
\begin{subfigure}[h]{0.42\linewidth}
\includegraphics[trim=0.5cm 0 0 0, clip, width=\linewidth]{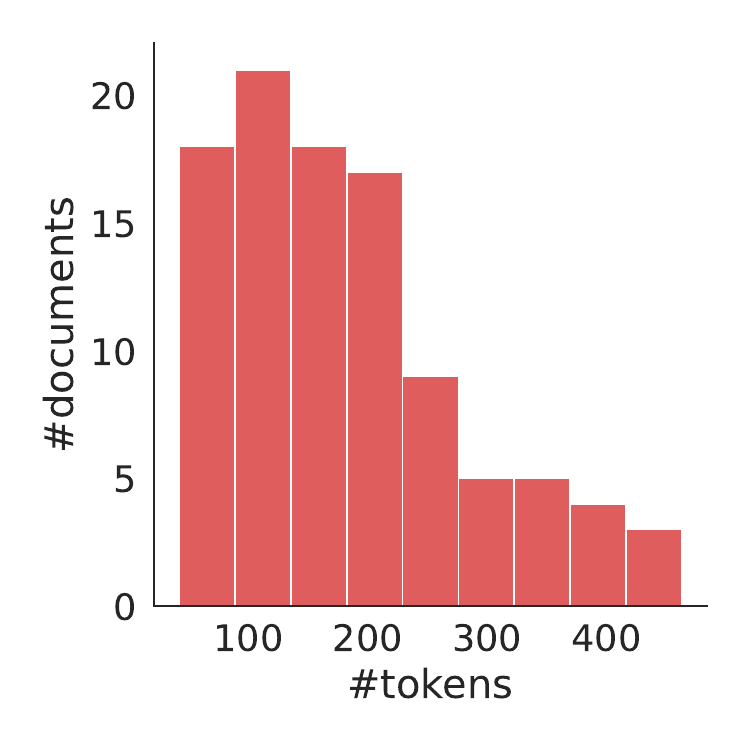}
\caption{French.}
\end{subfigure}
\vfill
\begin{subfigure}[h]{0.42\linewidth}
\includegraphics[trim=0 0 0.5cm 0, clip, width=\linewidth]{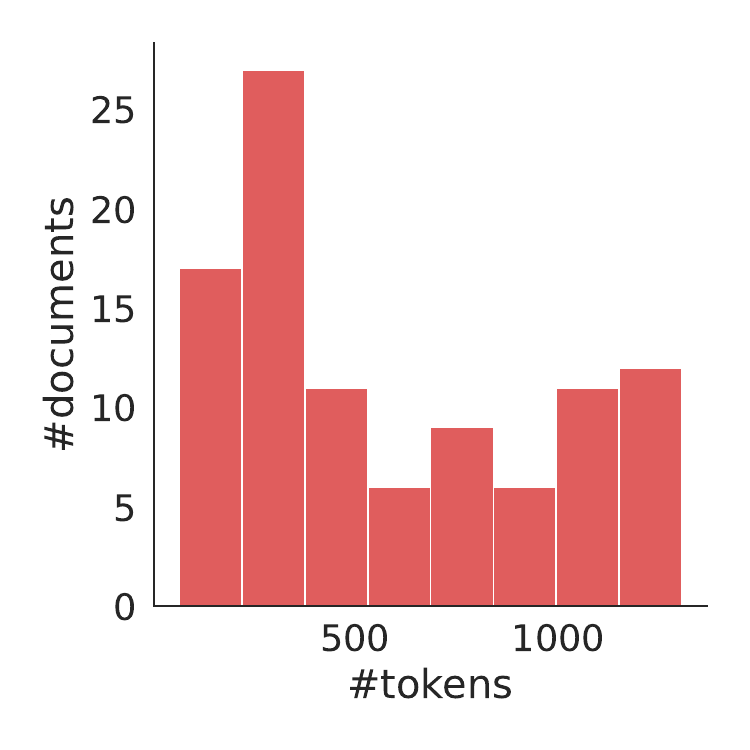}
\caption{Japanese.}
\end{subfigure}

\caption{The distribution of document length of the German (a), French (b), and Japanese (c) data using the number of tokens. Note the different scaling on the axes.}
\label{fig:text_len_distribution}
\end{figure}

\begin{figure}
\centering
\begin{subfigure}[h]{0.69\linewidth}
\includegraphics[trim=0 0 0.5cm 0, clip, width=\linewidth]{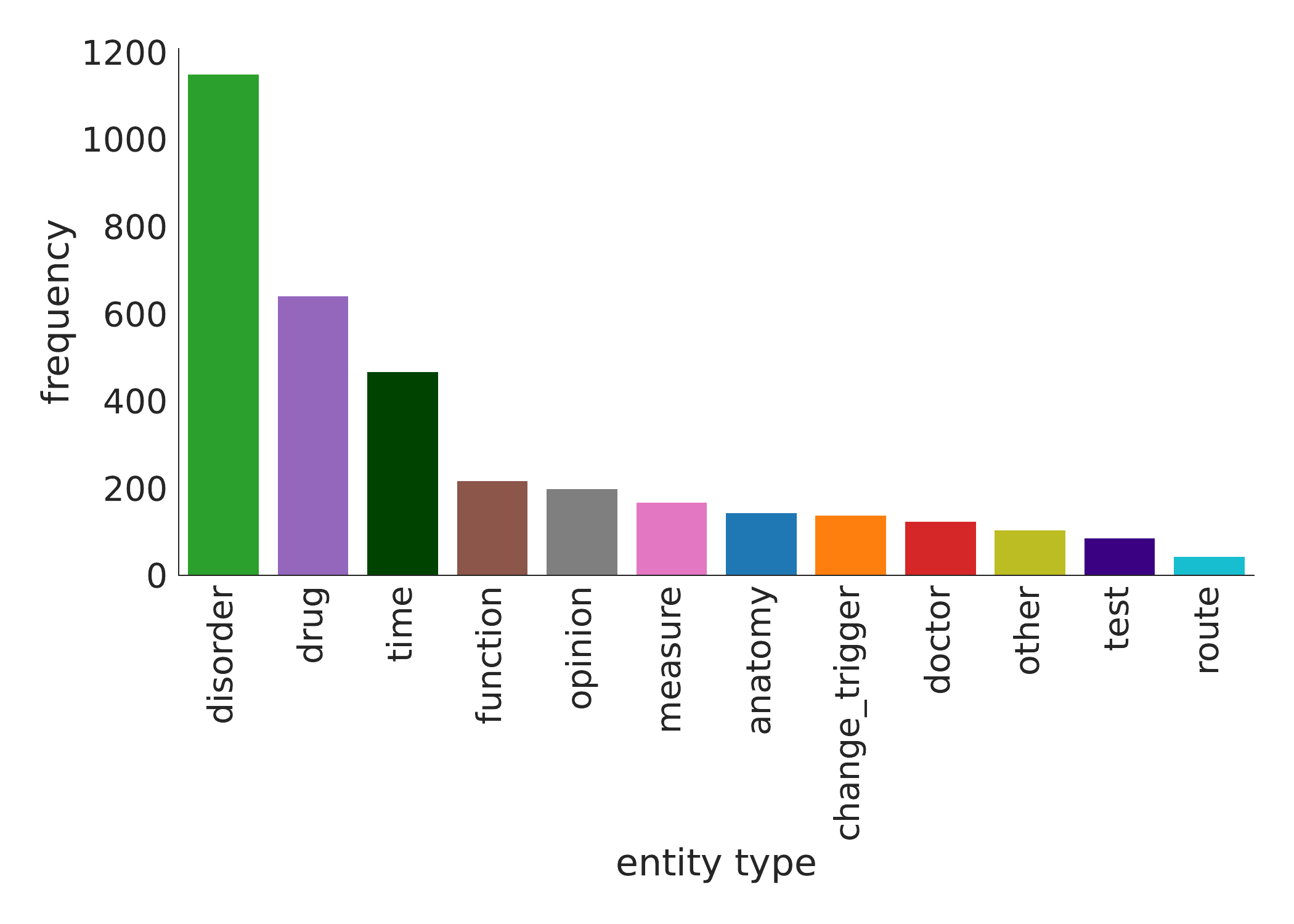}
\caption{German.}
\end{subfigure}%
\vfill
\begin{subfigure}[h]{0.69\linewidth}
\includegraphics[trim=0.5cm 0 0 0, clip, width=\linewidth]{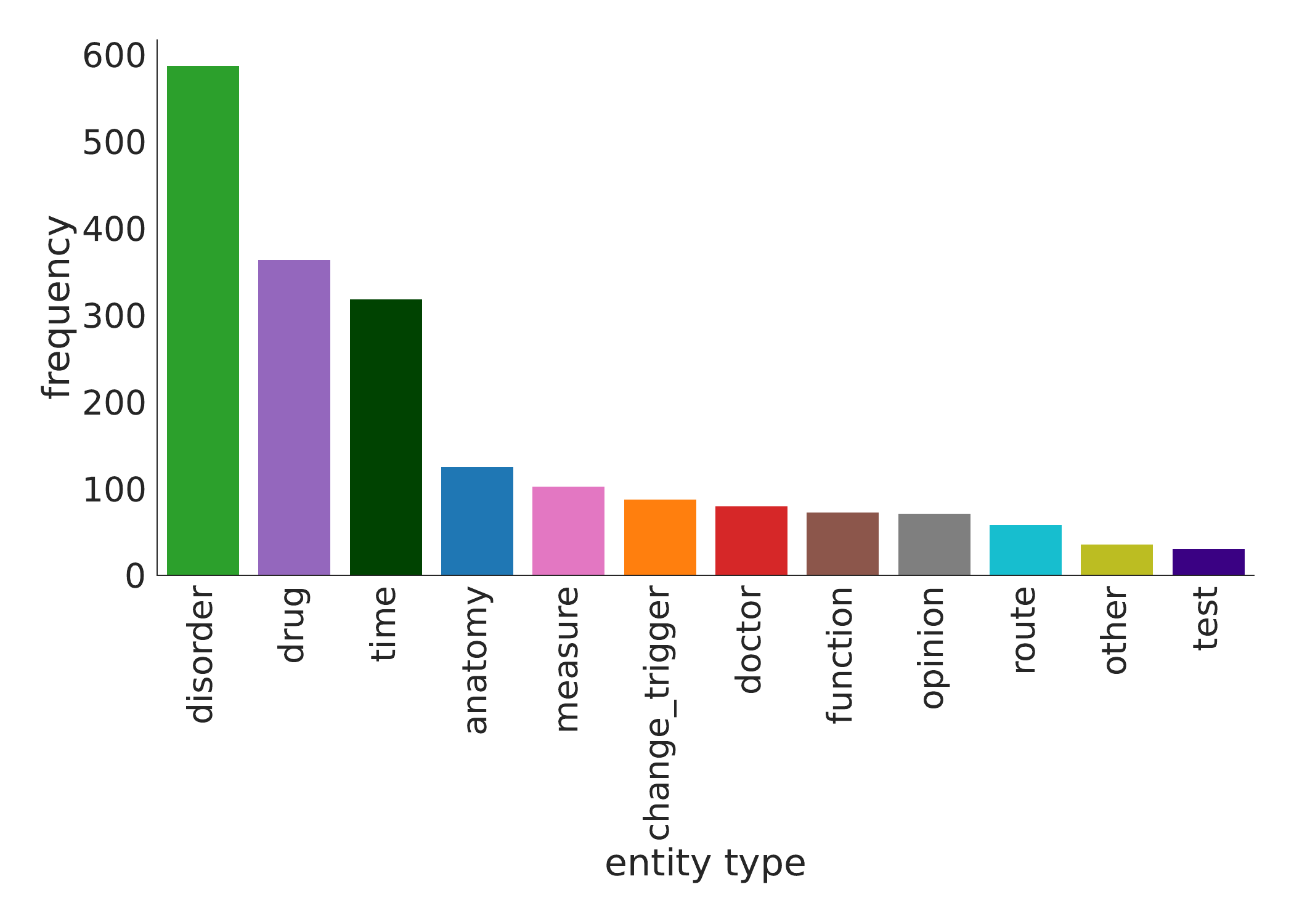}
\caption{French.}
\end{subfigure}%
\vfill
\begin{subfigure}[h]{0.69\linewidth}
\includegraphics[trim=0 0 0.5cm 0, clip, width=\linewidth]{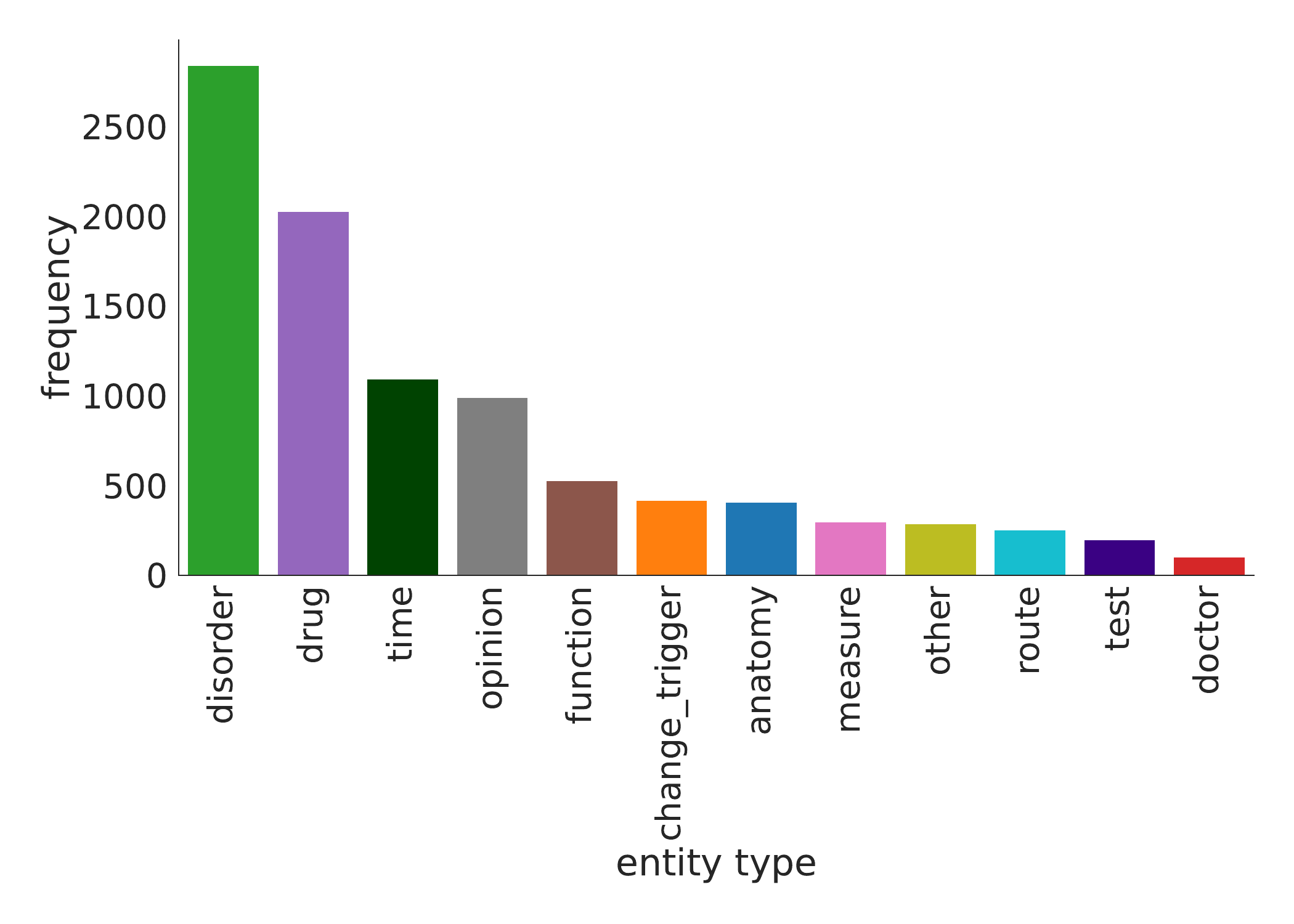}
\caption{Japanese.}
\end{subfigure}

\caption{The distribution of entity types across all documents for German (a), French (b), and Japanese (ja). Note the difference in scale when comparing the three languages. }
\label{fig:entity_distribution}
\end{figure}

\begin{figure}[h]
\centering
\begin{subfigure}[h]{0.60\linewidth}
\includegraphics[trim=0 0 0.5cm 0, clip, width=\linewidth]{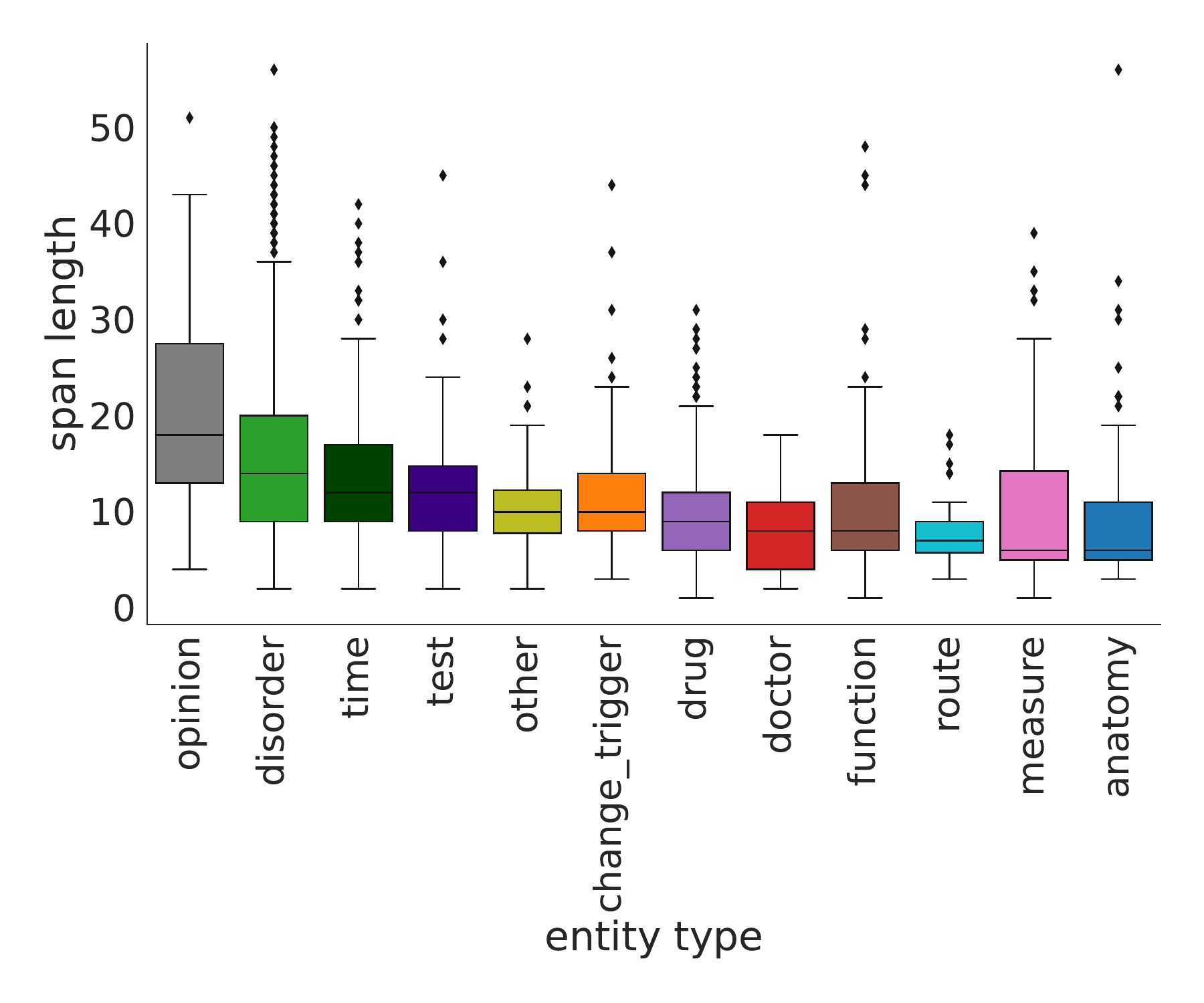}
\caption{German.}
\end{subfigure}
\vfill
\begin{subfigure}[h]{0.60\linewidth}
\includegraphics[trim=0.5cm 0 0 0, clip, width=\linewidth]{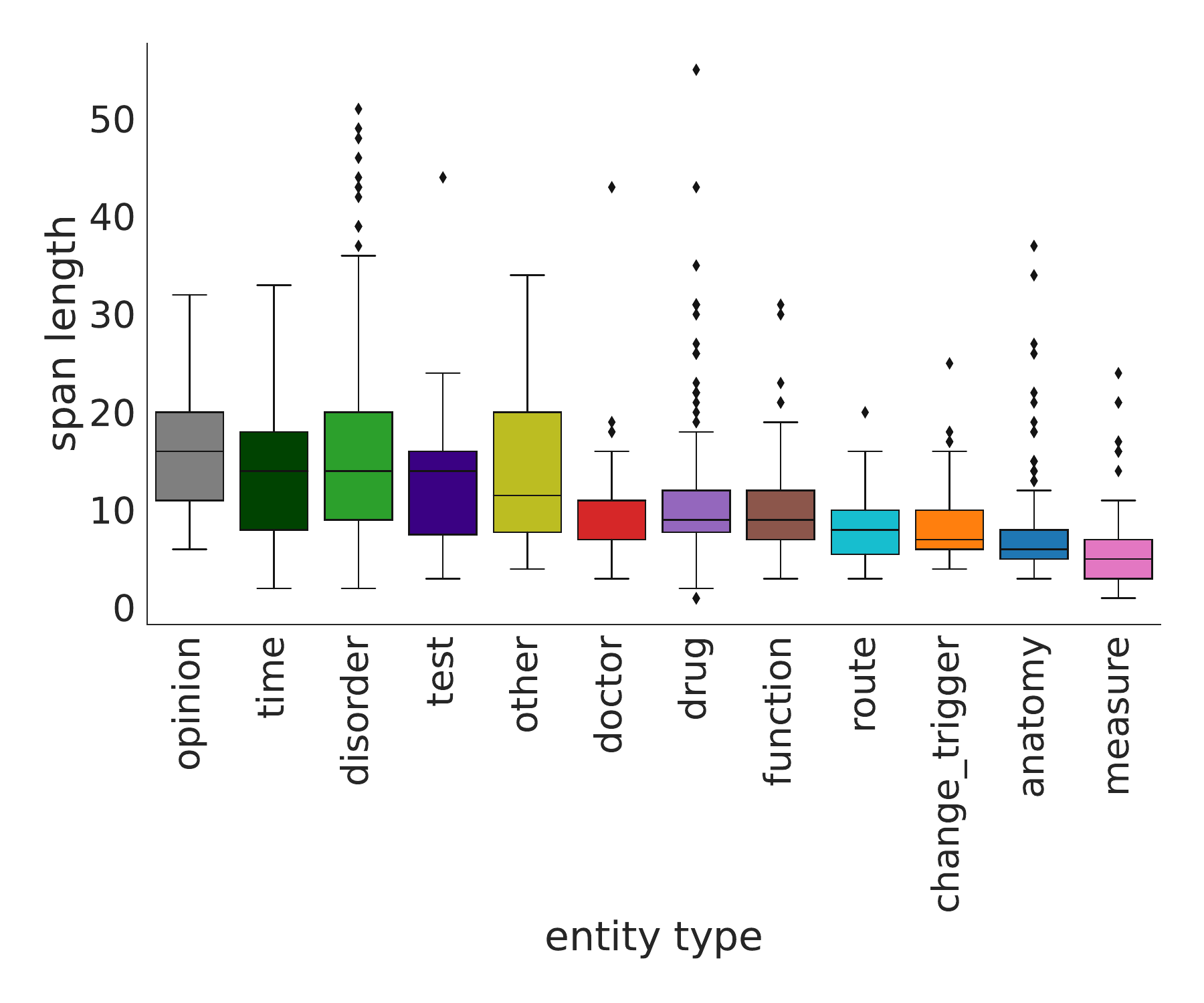}
\caption{French.}
\end{subfigure}
\vfill
\begin{subfigure}[h]{0.60\linewidth}
\includegraphics[trim=0 0 0.5cm 0, clip, width=\linewidth]{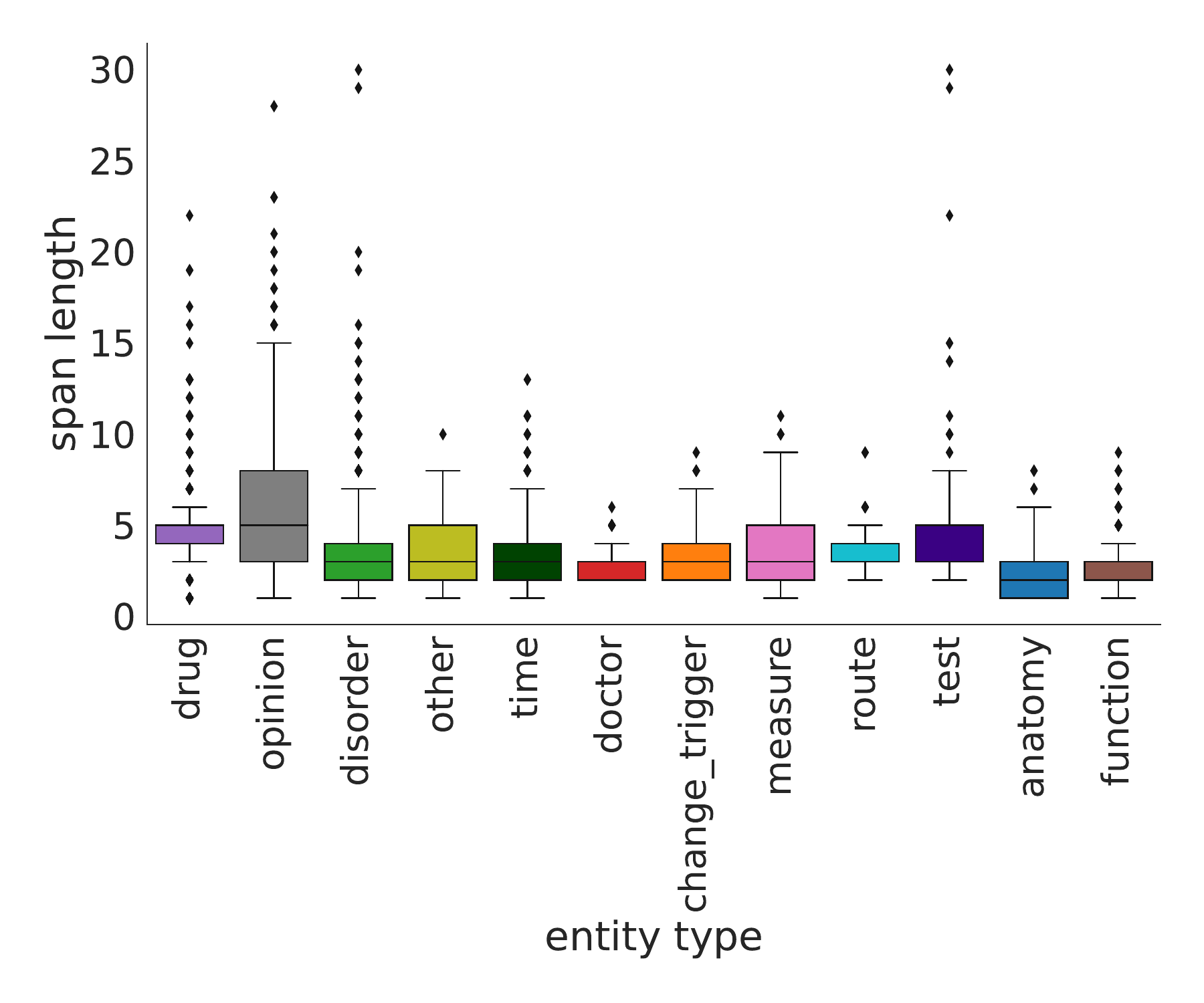}
\caption{Japanese.}
\end{subfigure}
\caption{The distribution of span length per entity type for German (a), French (b), and Japanese (c).}
\label{fig:entity_span_distribution}
\end{figure}

\begin{figure}[h]
\centering
\begin{subfigure}[h]{0.69\linewidth}
\includegraphics[trim=0 0 0.5cm 0, clip, width=\linewidth]{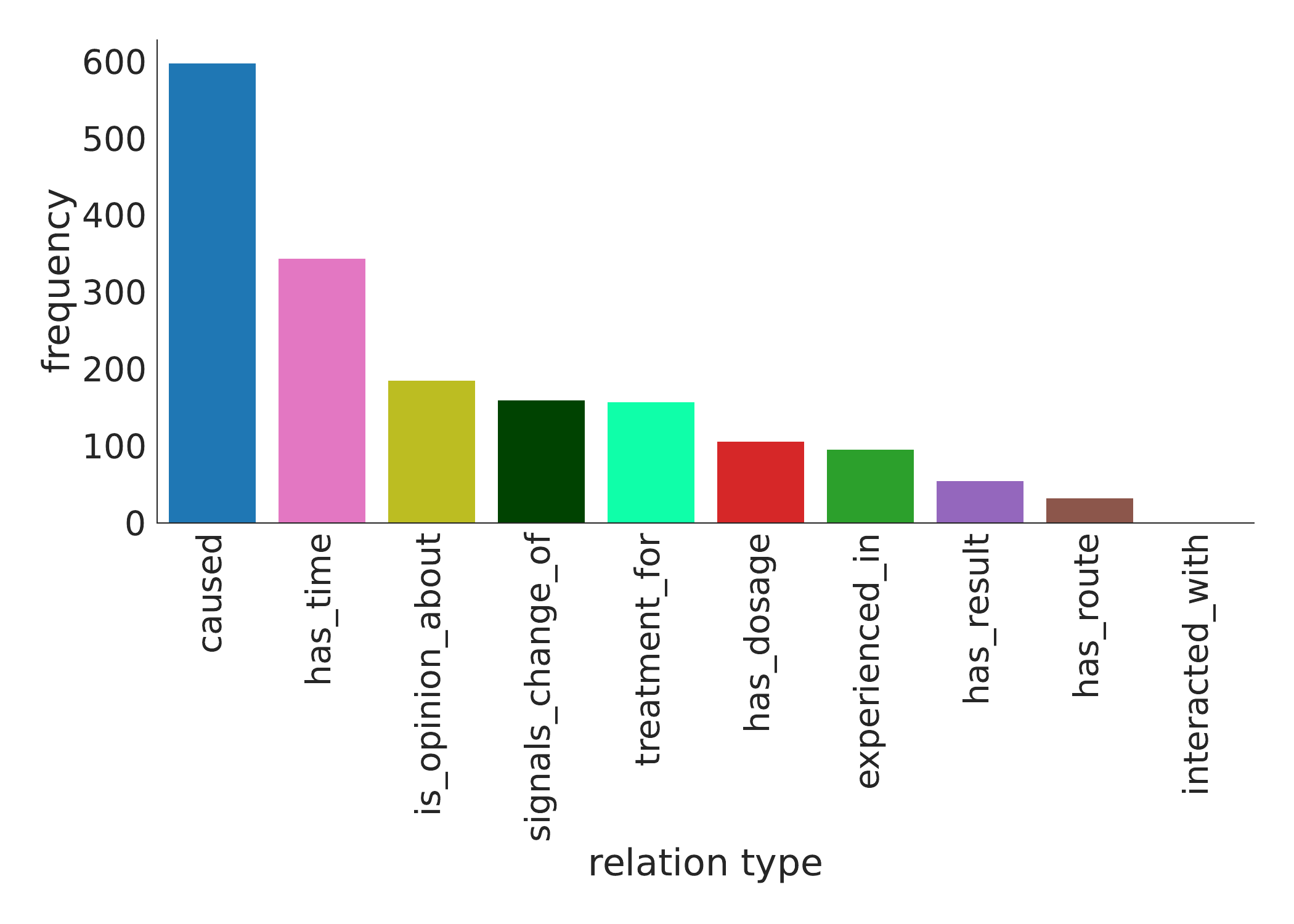}
\caption{German.}
\end{subfigure}
\vfill
\begin{subfigure}[h]{0.69\linewidth}
\includegraphics[trim=0.5cm 0 0 0, clip, width=\linewidth]{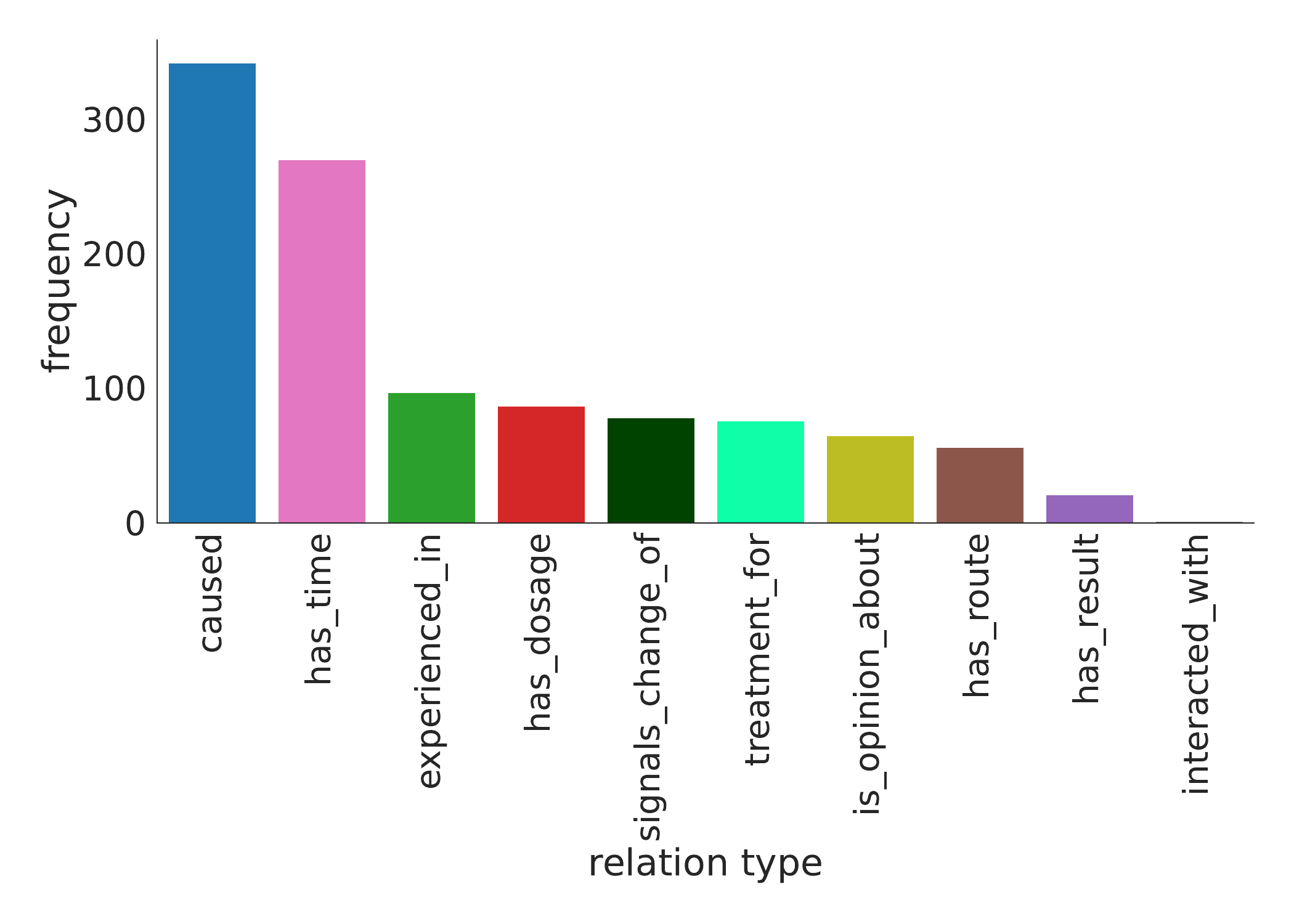}
\caption{French.}
\end{subfigure}%
\vfill
\begin{subfigure}[h]{0.69\linewidth}
\includegraphics[trim=0 0 0.5cm 0, clip, width=\linewidth]{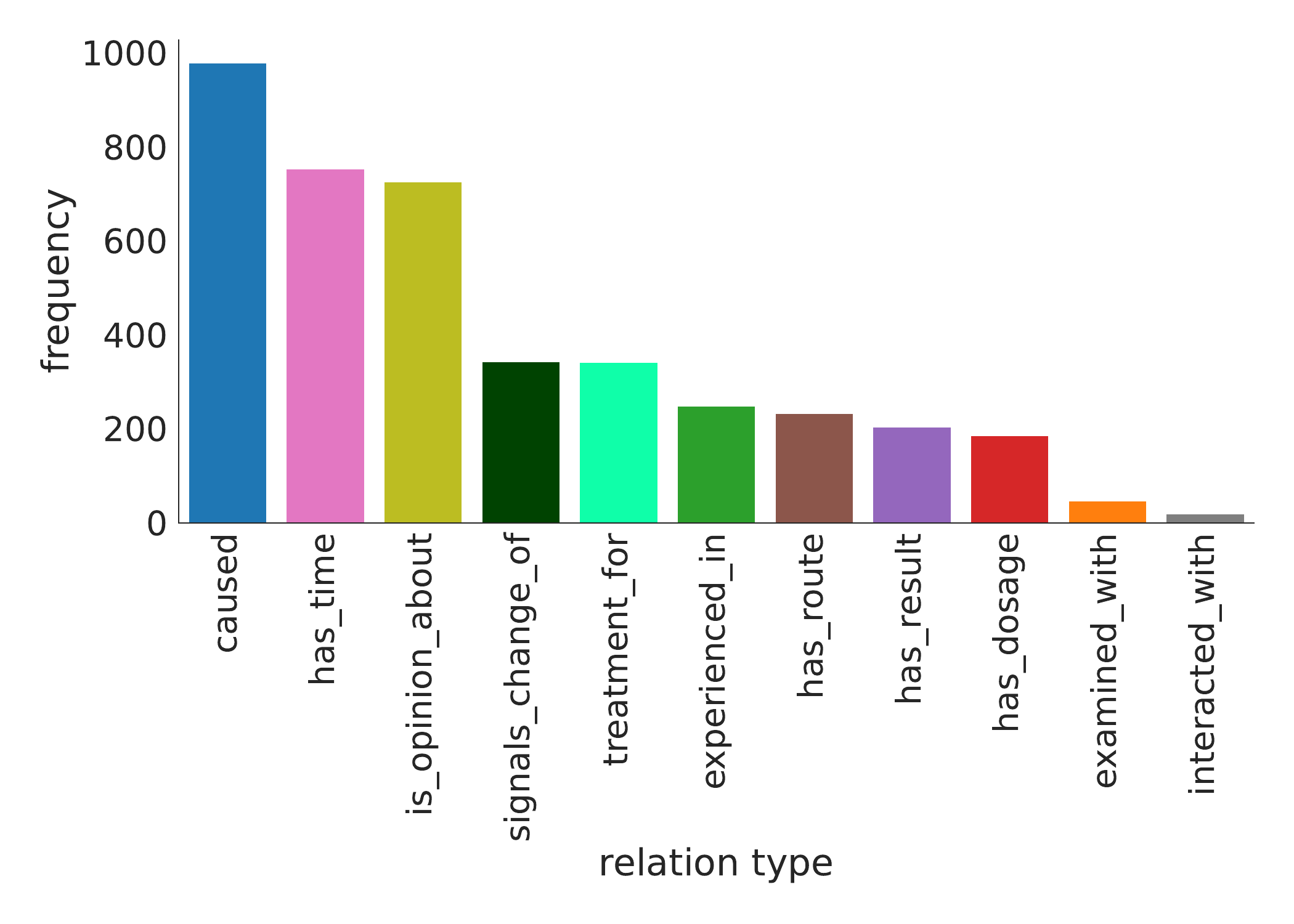}
\caption{Japanese.}
\end{subfigure}
\caption{The distribution of relation types for German (a), French (b), and Japanese (c). Note the difference in scale when comparing the three languages. }
\label{fig:relation_distribution}
\end{figure}

\begin{figure}[h]
\centering 

\begin{subfigure}[h]{0.68\linewidth}
\includegraphics[trim=0 0 0.5cm 0, clip, width=\linewidth]{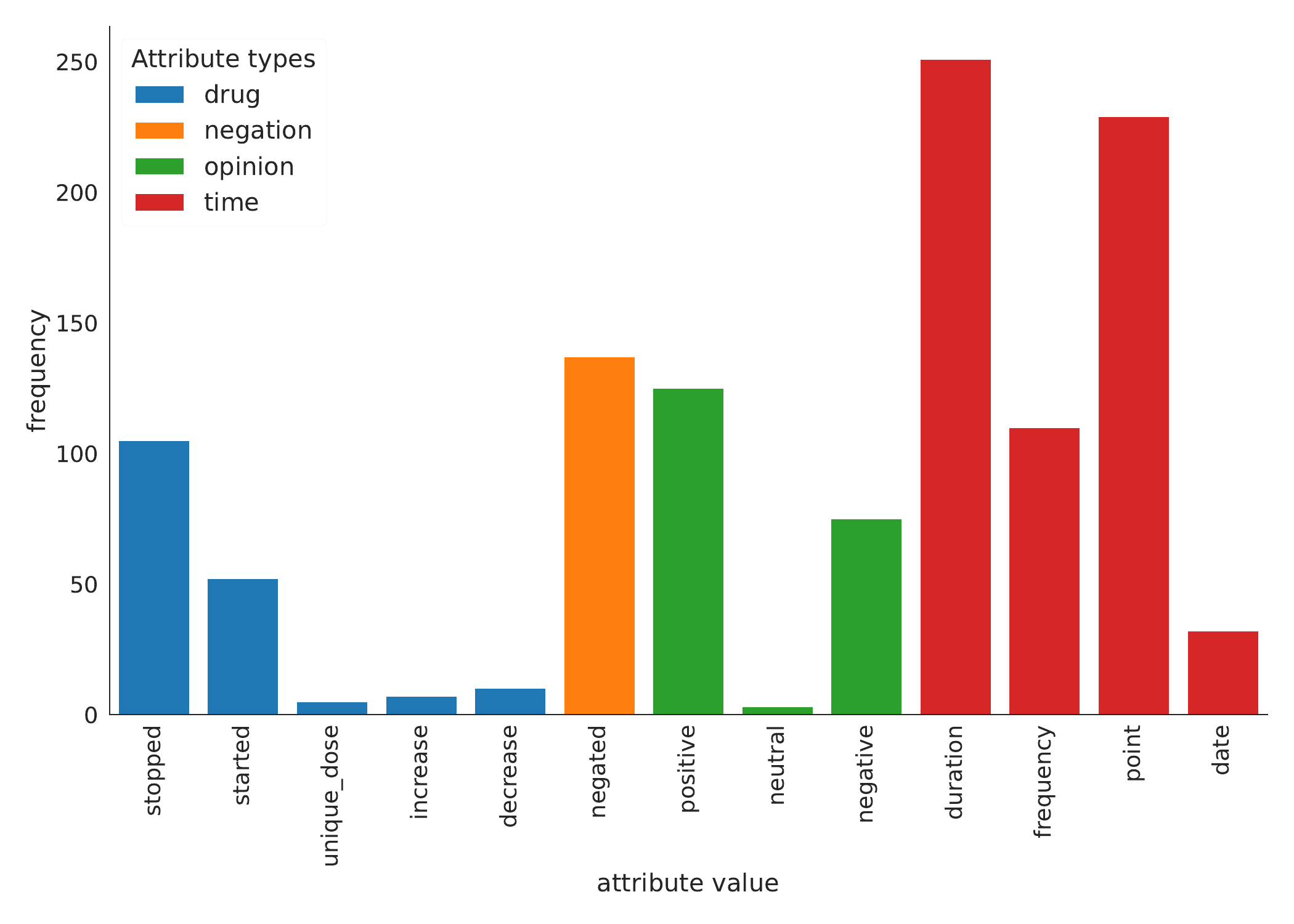}
\caption{German.}
\end{subfigure}
\vfill
\begin{subfigure}[h]{0.68\linewidth}
\includegraphics[trim=0.5cm 0 0 0, clip, width=\linewidth]{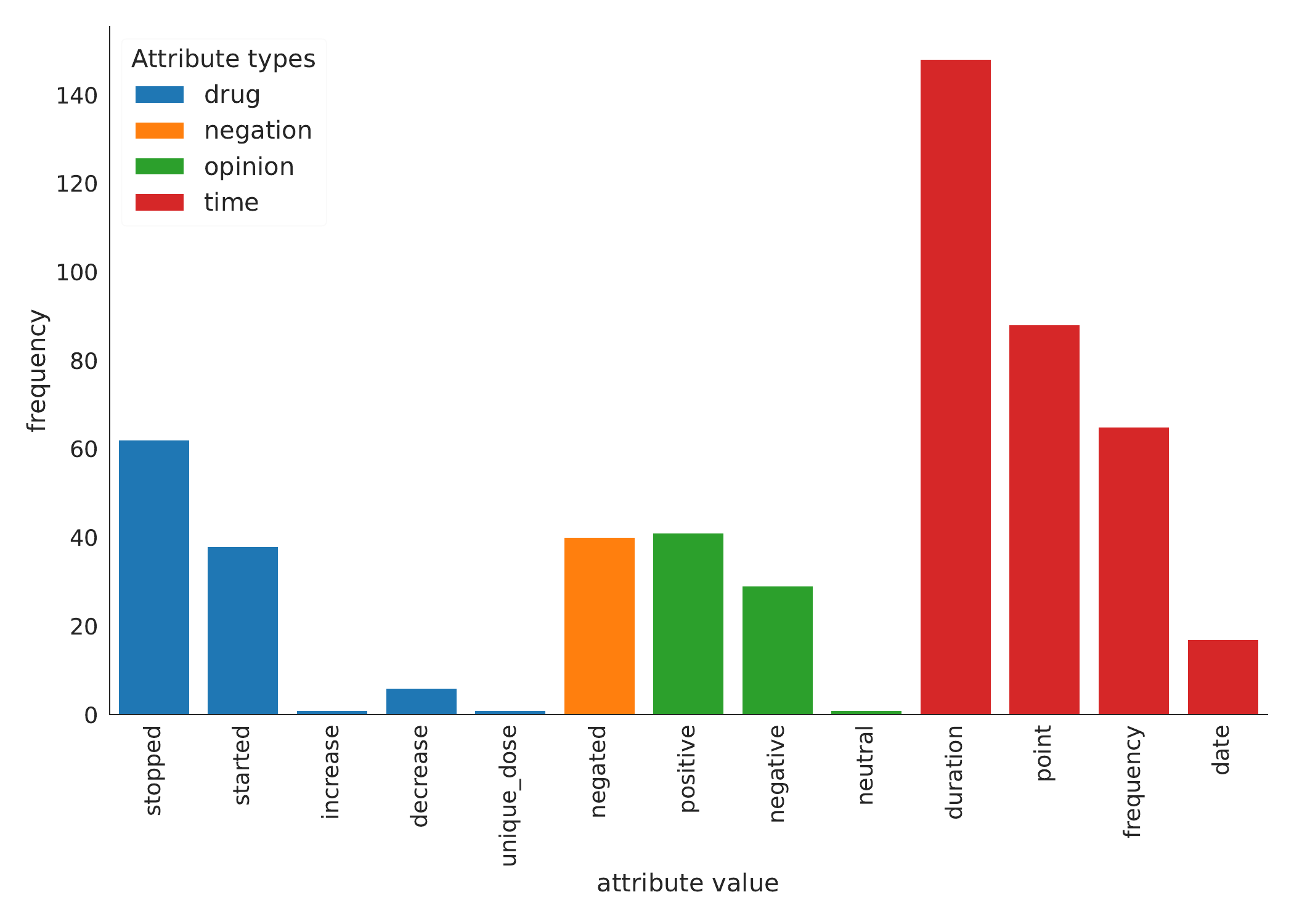}
\caption{French.}
\end{subfigure}%
\vfill
\begin{subfigure}[h]{0.68\linewidth}
\includegraphics[trim=0 0 0.5cm 0, clip, width=\linewidth]{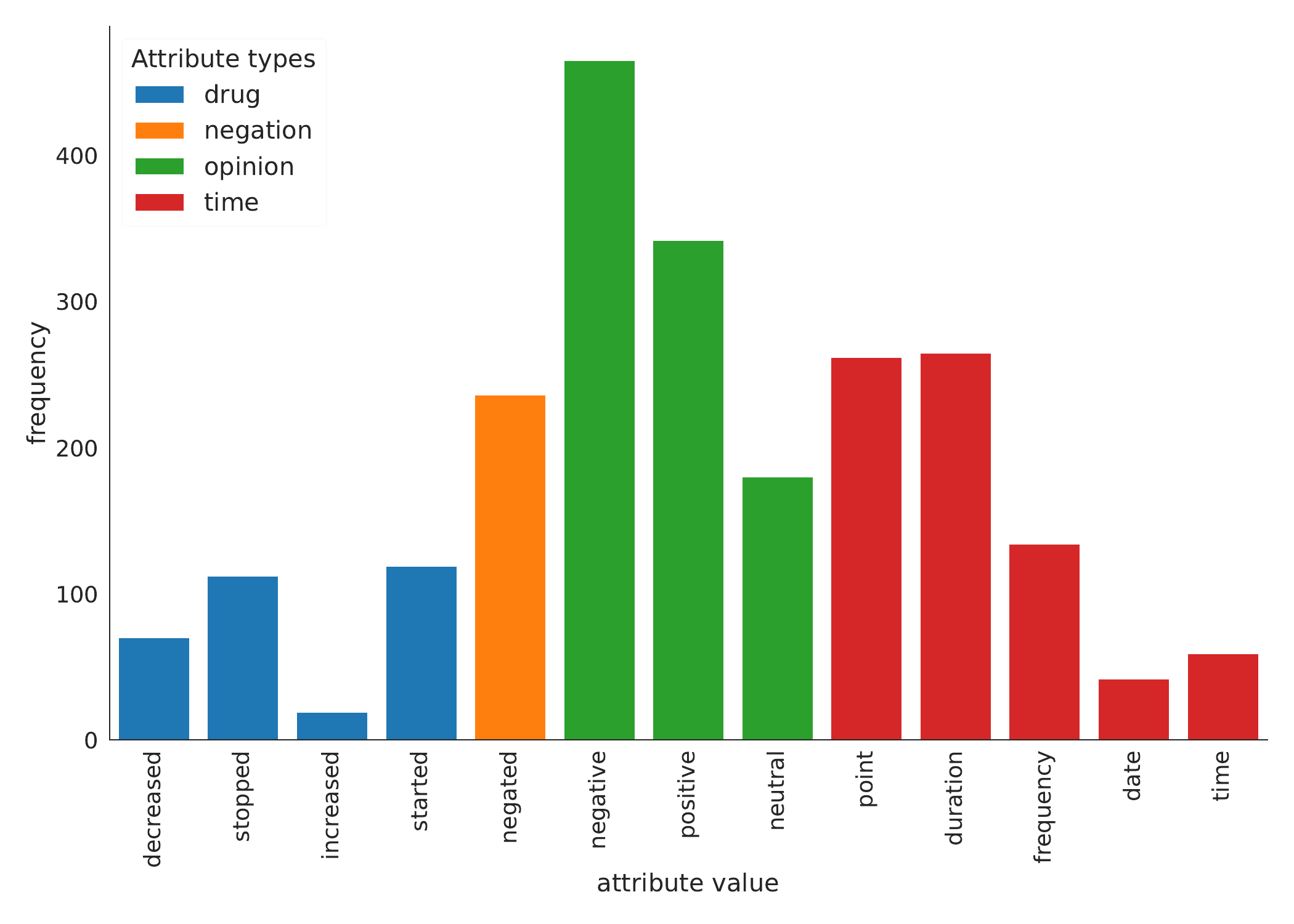}
\caption{Japanese.}
\end{subfigure}
\caption{The distribution of attribute values for each attribute type and for each language: \texttt{time} (duration, frequency, point in time, date), \texttt{opinion} (positive, neutral, negative), \texttt{drug} (stopped, started, unique dose, increase, decrease) and negation (only shown if an expression is negated).}
\label{fig:attribute_value_dist}
\end{figure}

\Cref{tab:overview_final_dataset} shows detailed statistics of the presented dataset. \Cref{tab:complicated_anno} shows the number of complex entity annotations.
We further show the distribution of document length in \Cref{fig:text_len_distribution}, the number of mentions per entity type in \Cref{fig:entity_distribution}, the span lengths per entity type in \Cref{fig:entity_span_distribution}, the number of relations per type in \Cref{fig:relation_distribution}, and the distribution of attribute values in \Cref{fig:attribute_value_dist}.

\begin{table}[h]
\centering
\begin{tabular}{@{}lrrrrrrrrr@{}}
\toprule
            & \multicolumn{1}{l}{} & \multicolumn{4}{c}{\textbf{\#tokens}}                        & \multicolumn{4}{c}{\textbf{\#sentences}}                       \\ \midrule
            & \textbf{\#docs}      & \textbf{total} & \textbf{mean} & \textbf{max} & \textbf{min} & \textbf{total} & \textbf{mean} & \textbf{max} & \textbf{min} \\ \midrule
\textbf{German} & 118                  & 29,032         & 246.03        & 815          & 55           & 1,674          & 14.19         & 50           & 1            \\
\textbf{French} & 100                  & 18,184         & 181.84        & 463          & 42           & 969            & 9.69          & 25           & 1            \\ 
\textbf{Japanese} & 99       & 58,024    & 586.10    & 1,303    &  68     & 2,165   & 21.87      &   58    &  4             \\ \bottomrule
\end{tabular}
\caption{An overview of the currently annotated data in German, French, and Japanese. It shows the number of documents for each language, the total number of tokens and sentences, and the mean, minimum, and maximum number of tokens and sentences per document. }
\label{tab:overview_final_dataset}
\end{table}

\begin{table*}[h]
    \centering
    \begin{tabular}{lcccc}
        \toprule
         & \textbf{\#Discontinuous} & \textbf{\#Overlapping} & \textbf{\#Total} \\
            
        \midrule
        \textbf{German} & 51 & 5 & 5346 \\
        \textbf{French} & 24 & 0 & 9454 \\
        \textbf{Japanese} & 25 & 10 & 9818 \\        
        \bottomrule
    \end{tabular}
    \caption{The number of discontinuous and overlapping entity annotations in German, French, and Japanese.}
    \label{tab:complicated_anno}

\end{table*}

\end{document}